\documentclass{article}
\usepackage{PRIMEarxiv}
\usepackage[numbers, sort&compress]{natbib}
\usepackage[inkscapelatex=false]{svg}
\usepackage[utf8]{inputenc}
\usepackage[T1]{fontenc}    
\usepackage[hidelinks]{hyperref}     
\usepackage{url}            
\usepackage{booktabs}       
\usepackage{amsfonts}       
\usepackage{nicefrac}       
\usepackage{microtype}      
\usepackage{lipsum}
\usepackage{fancyhdr}       
\usepackage{subcaption, graphicx}       

\usepackage{array}
\usepackage{enumitem}
\usepackage{lineno}

\title{Censoring chemical data to mitigate dual use risk}

\author{
  Quintina Campbell\\
  Department of Chemical Engineering\\
  University of Rochester\\
  Rochester, New York\\
  \texttt{qcampbe2@ur.rochester.edu} \\
    \And
  Jonathan Herington \\
  Department of Health Humanities \& Bioethics \\
  University of Rochester \\
  Rochester, New York\\
  \texttt{jonathan.herington@rochester.edu} \\ 
    \And
  Andrew D. White \\
  Department of Chemical Engineering \\
  University of Rochester \\
  Rochester, New York \\
  FutureHouse Inc. \\ 
  San Francisco, CA \\
  \texttt{andrew.white@rochester.edu} \\
}

\begin{document}
\maketitle
\begin{abstract}
Machine learning models have dual-use potential, potentially serving both beneficial and malicious purposes. The development of open-source models in chemistry has specifically surfaced dual-use concerns around toxicological data and chemical warfare agents. We discuss a chain risk framework identifying three misuse pathways and corresponding mitigation strategies: inference-level, model-level, and data-level. At the data level, we introduce a model-agnostic noising method to increase prediction error in specific desired regions (sensitive regions). Our results show that selective noise induces variance and attenuation bias, whereas simply omitting sensitive data fails to prevent extrapolation. These findings hold for both molecular feature multilayer perceptrons and graph neural networks. Thus, noising molecular structures can enable open sharing of potential dual-use molecular data.
\end{abstract}

\section{Introduction}

Machine learning in chemistry has become increasingly open and accessible to any individual with a computer and basic knowledge of coding \cite{butler2018machine, moosavi2020role, kim2021comprehensive, ramos2025review}. 
As with nearly every other novel technology, there are societal risks and opportunities. Easy-to-use ML tools, as beneficial as they can be, have a ``dual use'' as easy-to-use tools for doing harm. Following the nomenclature used in biology to identify ``dual use research of concern'' \cite{u.s._government_united_2014}, we term these harms the dual use risks of predictive chemistry (DURPC).

The ``dual use problem'' refers to the fact that scientific research ``has the potential to be used for harm as well as for good'' \cite{miller_ethical_2008}. While the vast majority of attention in the last two decades has been on dual-use risks generated by research in biology \cite{evans_great_2013, lipsitch_ethical_2014, selgelid_gain--function_2016, koblentz_novo_2017} and strong artificial intelligence \cite{bostrom_superintelligence:_2014, uncccw_report_2016},  chemistry may also pose risks. In chemistry, research investigating toxicity, flammability and some drug properties has caused concern as possible vectors for dual use risks \cite{tucker_innovation_2012, mehlich_chemistry_2018} and is the basis of long standing restrictions on the manufacture or possession of certain compounds or precursors \cite{noauthor_chemical_nodate, organization_for_the_prohibition_of_chemical_weapons_convention_1998}. 

Concern over the dual use risks posed by chemistry are being accelerated by machine learning. Cutting edge large language models (LLMs) can answer and solve graduate level scientific questions \cite{rein2023gpqa}, and can automate even more challenging tasks when they are incorporated in agent frameworks \cite{white2023assessment, bran2023chemcrow, mirza2024large}. 
These advanced capabilities have simultaneously raised safety concerns \cite{he2023control, tang2024prioritizing}, including concrete examples where LLMs have assisted in synthesizing dangerous compounds \cite{rose2022openai,bran2023chemcrow,boiko2023emergent}.
These risks are potentially compounded by open-weight models, which are not subject to active moderation of inputs and outputs. While open-weight LLMs can have safety features integrated into the model itself, others have demonstrated that these safety measures can be extracted or circumvented \cite{qi2023fine, gopal2023will}. Finally, visual language models (VLMs), should not be overlooked in the context of dual use risks, as it is possible that hazardous information, such as a recipe for nerve gas, can be passed in a form of images. Visual modality can bypass safety alignments of LLMs \cite{gong2023figstep}. The proliferation of different risks, mitigation strategies, and counter-strategies has made it difficult for institutions and individuals to identify appropriate mechanisms for controlling DURPC. 

\begin{figure}[ht]
    \centering
    \captionsetup{width=0.9\linewidth}
    \includegraphics[width=0.9\textwidth]{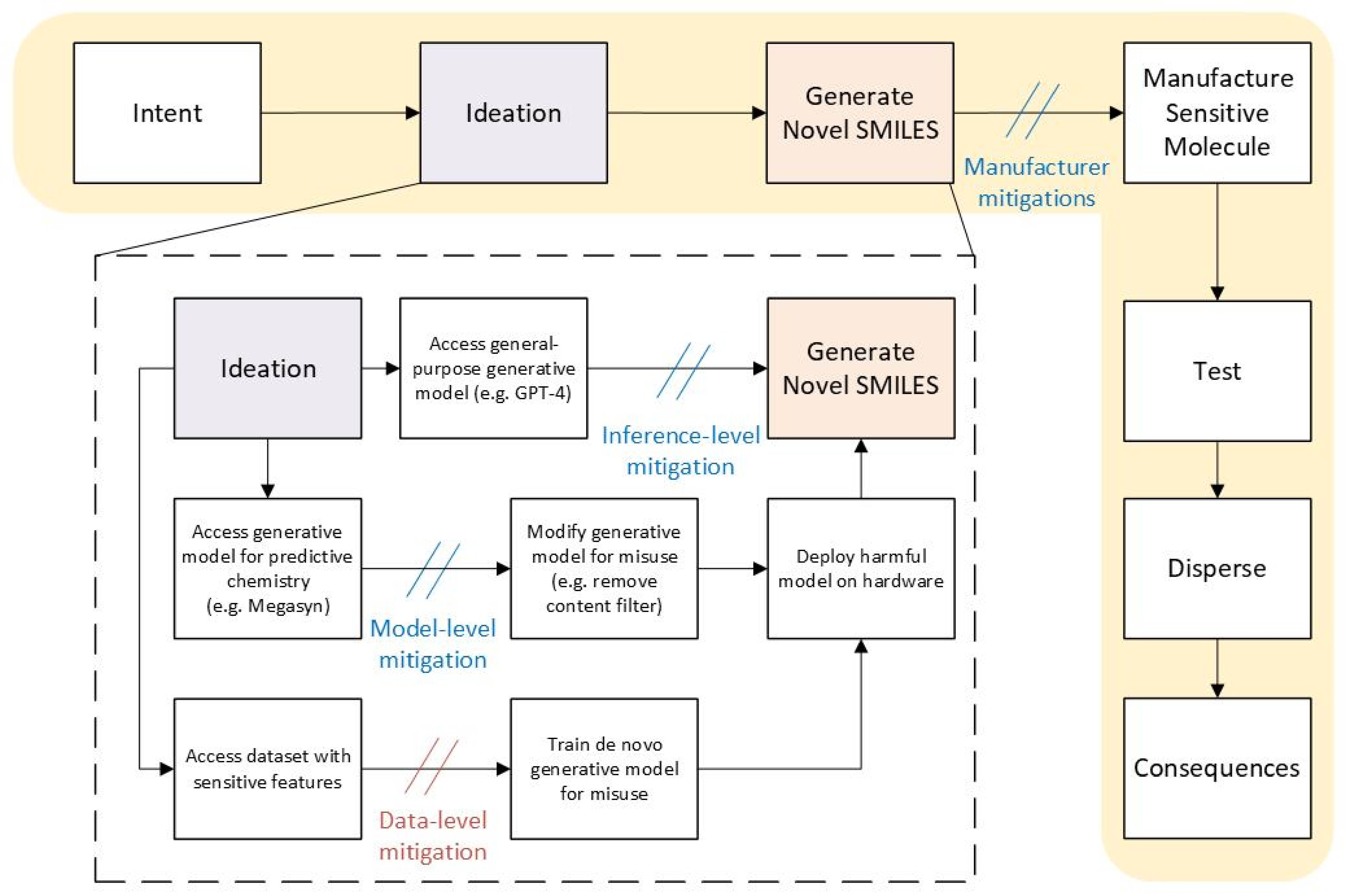}
    \caption{A chain risk framework for Dual Use Risks in Predictive Chemistry (DURPC). Starting with an actor with intent to do harm (top left), each barrier must be overcome for the consequences to be realized (bottom right). Three distinct pathways are provided for the step between ideation and generating a novel simplified molecular-input line-entry system (SMILES) structure. Points of intervention and general mitigation strategies are identified by cross-hatches. Our proposed intervention to reduce DURPC (i.e. data-level mitigation) is highlighted in red.}
    \label{fig:chain}
\end{figure}

In order to manage DURPC effectively, we develop a more detailed model of the risk landscape for computational chemistry and its ethical dimensions. Prior models for dual-use risks have tended to focus on the life sciences \cite{gryphon_scientific_risk_2015, tucker_innovation_2012, national_academies_of_sciences_engineering_and_medicine_biodefense_2018} and we extend a simple chain risk framework, previously used in the context of biological risk \cite{sandberg_who_2020}. On this model (see Figure \textbf{\ref{fig:chain}}), actors with intentions to cause harm must undertake multiple steps to realize their intent - from settling on a method of causing harm (i.e. explosives, toxins, biological agents, etc.), through design, manufacture, testing and deployment. Each step poses a barrier to an agent realizing harm, and overcoming each barrier requires a certain level of expertise or access to resources (i.e. ``agent power'' according to Sandberg and Nelson \cite{sandberg_who_2020}). This includes how easy it is to source precursors or materials, the level of expertise needed to use the technology maliciously, and the mechanisms for deploying resulting weapons. By estimating the likelihood of different kinds of agents (i.e. individuals, small organization, or states) overcoming different barriers, this model helps estimate the effect of different policy interventions for DURPC.

In this paper, we focus on mitigation strategies to prevent the design of novel chemical structures that pose dual use risks. We distinguish between low, moderate and high resource agents in order to help characterize different pathways towards generation of novel chemical structures using generative models. We characterize \textit{Low resource agents} as ordinary laypeople without a sophisticated understanding of machine learning or chemical synthesis, and without access to resources beyond those available in the consumer retail market. \textit{Moderate resource agents} are committed, reasonably knowledgeable individuals or small groups with moderate access to resources (i.e., lab equipment or high performance computing resources) that may require knowledge of specialized technical retailers in order to synthesize novel compounds (i.e., chemical synthesis companies). A classic example of such groups is the Aum Shinrikyo cult, which manufactured and released Sarin into the Tokyo subway in 1995 \cite{cameron_multi-track_1999}. \textit{High-resource agents} are so-called "state-level actors": committed organizations with the scope to recruit subject-matter experts, gather equipment or precursors on an industrial scale, and the time and discipline to engage in multi-year research and development. The risks posed by this last category may be particularly difficult to control, since state-level actors may be able to back-engineer or reproduce their own datasets and models and thus circumvent technical efforts to prevent misuse. Our goal is to limit DURPC for low to moderate resource agents.

A final consideration is the ethical trade-offs involved in any intervention to limit DURPC \cite{selgelid_gain--function_2016}. In general, the scientific community wants to make toxicity data generally available, so that researchers and practitioners can generate their own predictors to generate beneficial knowledge about novel molecules. This position aligns with scientific norms of openness and reproducibility, which have historically accelerated innovation and safety improvements in chemistry. Some philosophers \cite{evans_great_2013, kuhlau_precautionary_2011} of science have argued that traditional scientific norms may require recalibration towards "precautionary" approaches when technologies offer unprecedented potential for harm alongside uncertain benefits. On the other hand, precautionary approaches can be self-defeating if they prevent research which defends against risks other than DURPC which may be more likely or more harmful \cite{clarke_precautionary_2013}. Maintaining the accuracy of toxicity predictors is critical to mitigating unintentional harm from generative models which identify novel, potentially useful chemical compounds. In this respect, the effect of DURPC interventions on the overall accuracy of predictors, or their accuracy with respect to different zones of chemical space, must be carefully evaluated.

\begin{figure}[ht]
    \centering
    \captionsetup{width=0.9\linewidth}
    \includegraphics[width=0.8\textwidth]{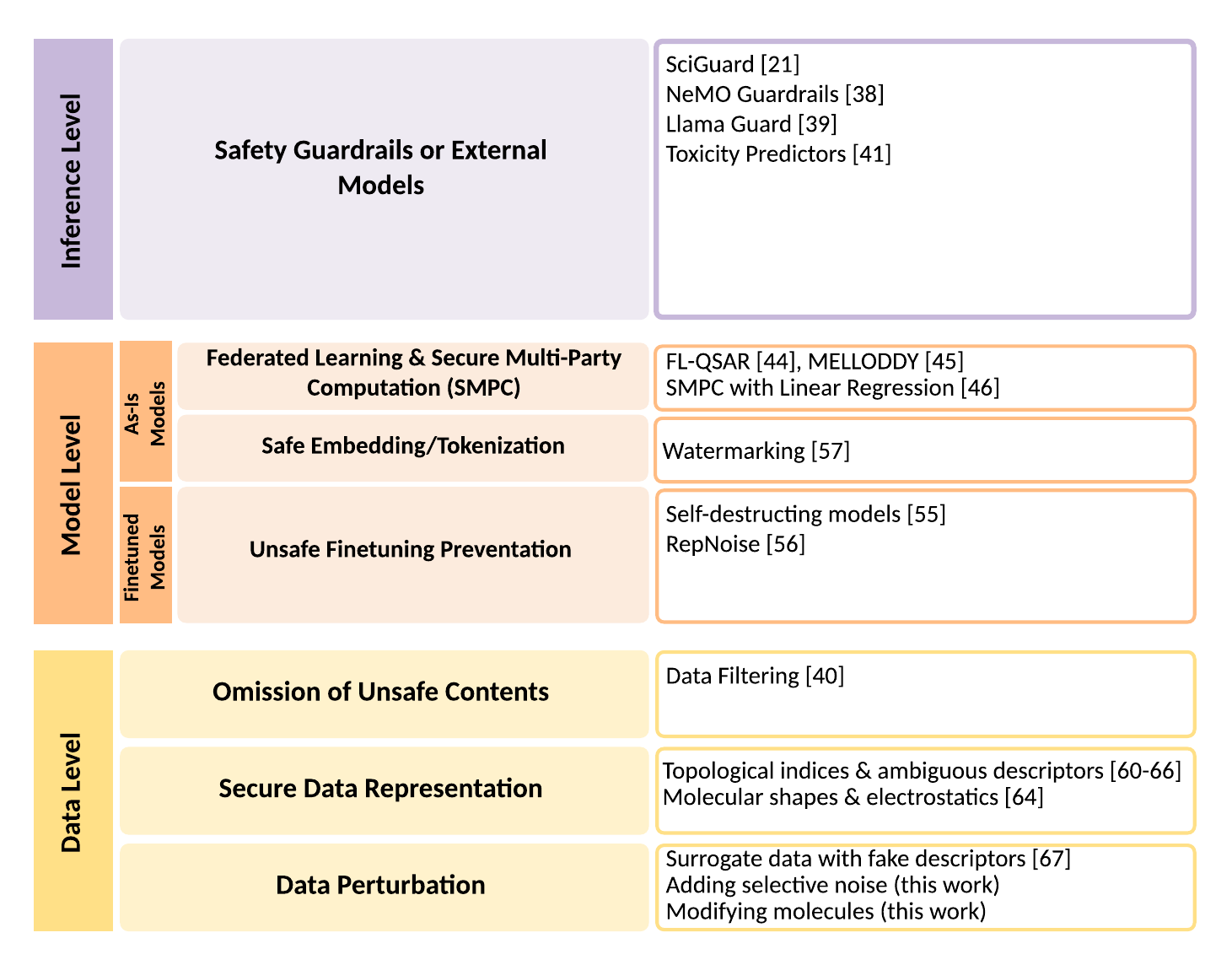}
    \caption{Summary of strategies for mitigating DURPC, categorized into three levels based on where the strategy is implemented: dataset, model, and inference.}
    \label{fig:risk_mitigations}
\end{figure}

\section{Related Works}
Mitigating DURPC shares a similar basic structure as ensuring privacy and safety in machine learning and artificial intelligence (AI): it involves identifying classes of sensitive information and generating strategies to effectively limit the disclosure of such information to end users. In the context of general AI privacy and safety, the classes of sensitive data might include demographic information, socioeconomic status, hate speech, criminal planning, to list a few \cite{yuan2024seval}.
Meanwhile, in chemistry, classes of sensitive information would consist of toxicity, highly energetic materials, nuclear or biological weapon precursors, and psychoactivity (narcotics), often categorized within the ``chemical, biological, radiological, nuclear'' (CBRN) hazard category \cite{phuong2024evaluating, tedeschi2024alert}, and sometimes with high yield ``explosives'' (CBRNE) \cite{vidgen2024introducing}. Strategies to counteract and mitigate dual usage of this information can be broadly categorized into three approaches: inference-level mitigation, model-level mitigation, and data-level mitigation (Figure \textbf{\ref{fig:risk_mitigations}}). 

\paragraph{Inference-level mitigation}
Inference-level methods to mitigate dual use risks, also known as system-level mitigations, are front line defenses and often consist of external safety guardrails \cite{rebedea-etal-2023-nemo, inan2023llama, grattafiori2024llama3} or models \cite{he2023control, challacombe2024towards}. These external techniques would detect and control potential harm in user input, model output, or both. In particular, Llama Guard \cite{inan2023llama} can detect inputs or outputs that facilitate the construction of illegal weapons and regulated substances. An external LLM agent SciGuard \cite{he2023control} is designed to detect and control sensitive information in scientific models. The limitations with the strategies above are that they do not mitigate the risks posed by open-source models, and are prone to be bypassed by medium and high resource agents by jailbreaking or similar.

\paragraph{Model-level mitigation}
Model-level mitigations can be divided into two subcategories based on their objectives: 1) model training or architecture designated to secure sensitive data or model weights, and 2) safety alignment of models to lower the probability of unsafe outputs in the first place. To secure data, federated learning is commonly used and allows data owners to collaboratively train a model by aggregating their model weights into a global model without sharing their datasets \cite{mcmahan2017communication, yang2019federated}.
FL-QSAR and MELLODDY are few instances of federal learning models in science \cite{chen2020fl, heyndrickx2023melloddy}. Another similar method is to use secure multi-party computation (SMPC/MPC), which has been used for linear regressions with chemical data \cite{karr2005secure}. The limitations of federated learning and MPC are that while they may secure chemically sensitive information classes in data, they do not align models away from the sensitive regions in the chemical space.

For model-level alignment of AI, reinforcement learning with human feedback (RLHF) or other forms of reinforcement learning with safety cost functions are commonly used to reduce safety risks \cite{chennakesavalu2024energy, ouyang2022training, bai2022training, saisubramanian2022avoiding, zhang2018minimax, low2024safe, stooke2020responsive}. RLHF models become more challenging to red-team as they scale, whereas other model types show flat trends when scaled up \cite{ganguli2022red}. 
To prevent harmful fine-tuning, some defense mechanisms are ``self-destructing'' models to slow the convergence of harmful training \cite{mitchell2022selfdestructing} and representation noising, i.e. RepNoise \cite{rosati2024representation}, to reduce and remove harmful representations within models while fine-tuning. Additionally, safe embeddings or tokenization, like the 'watermarking' framework \cite{kirchenbauer2023watermark}, can tailor the model away from generating unsafe contents. However, high computational costs are a common limitation when utilizing reinforcement learning or other fine-tuning approaches to aligning models for chemical safety, or AI safety in general.

\paragraph{Data-level mitigation}
Finally, mitigating DURPC at the data level can be achieved through omission of sensitive data (also called data filtering) or data obfuscation. Data filtering, commonly used in large language models \cite{grattafiori2024llama3},
removes bias and unsafe content from training data to prevent introducing sensitive information to models. However, this is challenging in regression tasks for chemical data, where models can extrapolate to novel, potentially harmful chemicals, as seen in bioactivity models \cite{urbina_dual_2022}. 

Data obfuscation can be done by using ambiguous representation, surrogate data, or data perturbation.
Research on censoring chemical data dates back to a 2005 ACS symposium, which discussed exchanging chemical data without revealing molecular structures \cite{karr2005secure, filimonov2005relevant, masek2008sharing, faulon2005reverse, balaban2005can, varnek2005substructural, nicholls2005molecular, matlock2014sharing,  bologa2005descriptor, tetko2005surrogate, bradley2005share, trepalin2005centroidal, kaiser2005similarity, clement2005possibilities, swamidass2015securely}. While initially aimed at protecting intellectual property, this work can also apply to mitigating safety risks associated with revealing chemical structures. A key finding was that even minimal chemical information can expose structures, which led to reluctance to share pharmaceutical data \cite{filimonov2005relevant}. For example, molecular masses can be sufficient to infer structures \cite{masek2008sharing}. Later studies explored which descriptors could or could not reveal structures \cite{masek2008sharing}

Some studies propose ambiguous representations with high degeneracy (multiple molecules per representation) to censor chemical structures. Examples include topological indices \cite{faulon2005reverse, balaban2005can}, substructural fragment matrices \cite{varnek2005substructural}, molecular shape or electrostatics descriptors \cite{nicholls2005molecular}, chemical relationships \cite{matlock2014sharing}, and identical descriptors for distinct molecules to cause collision and confusion \cite{bologa2005descriptor}. However, these methods do not steer a model away from sensitive classes and thus possibly fail to prevent it from predicting sensitive classes. Furthermore, a study shows that reverse engineering a structure from a single descriptor is possible, suggesting that descriptors may not be reliable methods for censoring sensitive information and suggesting surrogate data as an alternative \cite{tetko2005surrogate}. This surrogate data replace the original molecule's fingerprints with those of similar molecules to mask chemical structures while maintaining model performance.

\subsection{Assessing Dual Use Risks}
While this work solely focuses on mitigating dual use risks, another equally important aspect of AI safety is assessing dual use risks. Most prior work has focused on identifying and evaluating these risks, even more in LLMs, commonly performed by red teaming \cite{ganguli2022red, mouton2024operational, patwardhan2024building, ge2023mart} and benchmarking \cite{yuan2024seval, li2024red}. A risk category containing CBRN(E) is included in a taxonomy for red teaming \cite{phuong2024evaluating} and benchmarks \cite{vidgen2024introducing, tedeschi2024alert, li2024wmdp, he2023control}. Compared to risk assessment, there is only a sparse literature focused on developing techniques to mitigate DURPC in machine learning and artificial intelligence. Nonetheless, many frameworks for AI safety propose multifaceted approaches rather than implementing a single technique \cite{tang2024prioritizing, barrett2024benchmark}, and the same is likely to be true for mitigating DURPC in machine learning and artificial intelligence. 

\subsection{Our Work: Adding Selective Noise to the Chemical Data}
Machine learning models are sensitive to dataset distribution, imbalance, and noise. As a method for mitigating DURPC, we propose selectively adding noise to specific portions of chemical datasets. This data-level mitigation approach adds noise to either the labels or features of only those data classes that are flagged as sensitive, such as highly flammable chemicals, toxins, or nerve agents. This method of selectively noising data points in the dataset can effectively contribute to mitigating dual usage for two main reasons. 

Firstly, we apply selective noise in datasets as a step toward enabling data sharing. As demonstrated by the bioactivity model \cite{urbina_dual_2022}, removing sensitive data points does not prevent the model from making sensitive predictions. Data collection is costly in terms of time and resources, and data scarcity is a major challenge in building machine learning models for science \cite{xu2023small}. This has led to growing interest in low-shot, zero-shot, and other small-data learning approaches \cite{altae2017low, wang2018lowshot}. When data can be shared safely, researchers gain access to valuable datasets while minimizing DURPC, thus fostering open collaboration and the advancement of chemical and biological technologies.

Secondly, this method provides a model-agnostic approach to mitigate the risks of DURPC. With increasing accessibility of machine learning tools and resources, individuals can build their own models. They can use LLMs to generate code to train a model \cite{white2023assessment}. Inference-level and model-level strategies may limit misuse of existing models but are insufficient for addressing potential dual use risks from custom models developed by individuals with access to sensitive data. Data-level mitigation addresses dual use concerns regardless of which architectures are being used. 

\section{Theory}\label{sect:theory}

We can write a data generating function (e.g. measuring a chemical property) as: $ f\left(\vec{x}\right) = y$, where $\vec{x}$ is a featurization of the molecule and $y$ is the measured property.\footnote{for simplicity we ignore inherent noise in labels} 
In the usual machine learning setting, we receive data $\mathcal{D}$ that are pairs of $(\vec{x}_i, y_i)$ from $f(\vec{x})$ to fit an approximation: $\hat{f}(\vec{x}; \theta)$. That approximation is defined by parameters $\theta$, which could be, for example, neural network weights or best-fit linear coefficients.

We can denote the sensitive regions via some function $s(y) = 1$ or $0$, like $y > y_t$, where $y_t$ denotes some threshold for label values to be sensitive. For simplicity, we consider only two classes of data: sensitive and non-sensitive, with $y_t$ representing a constant threshold used to distinguish between the two. The generalization error (expected mean squared error in the test data) is written as
$E_{x, \theta}[(\hat{f} - y(\vec{x}))^2]$,
where the expectation is taken over both the test point ($x$) and the parameters ($\theta$), which depend on $\mathcal{D}$. 

Our goal is that:
\begin{equation}
\label{eq:goal}
    E_{\vec{x},\theta|s = 1}\left[\left( \hat{f}(\vec{x};\theta) - y\right)^2\right] >> E_{\vec{x},\theta|s = 0}\left[\left(\hat{f}(\vec{x};\theta) - y\right)^2\right]
\end{equation}

where $s = 0$ means $s(y) = 0$. A baseline approach is to omit the points where $s(y) = 1$. For example, when predicting toxicity, the most toxic molecules considered sensitive would be removed from the training data. A clear drawback is that there is no control over the generalization error; we cannot prevent a model from learning via extrapolation.

Here we propose to generate noisy data, i.e. perturbed data, $\mathcal{D}' = \left(\vec{x}_i + \delta \vec{x}(y_i), y_i + \delta y(y_i)\right)$, where $\delta \vec{x}(y_i)$ and $\delta y(y_i)$ are controllable probability distributions and are dependent on whether the label $y_i$ resides in the sensitive region. These distributions should be chosen to minimize generalization error in our non-sensitive region and maximize in our sensitive region. Among noise types previously evaluated, zero-mean Gaussian noise is reported to be particularly effective at reducing deep learning model performance \cite{boonprong2018classification}.

A direct analysis of Equation~\ref{eq:goal} depends on the training procedure: $P(\theta|\mathcal{D}')$ \cite{adlam2020understanding}. Some intuition can be found by assuming that $\theta$ could be separated into $s=1$ and $s=0$ regions and then using the bias-variance decomposition of the generalization error \cite{geman1992neural}:

\begin{equation}
\label{eq:bias-var-decomp}
E_\theta\left[\left(\hat{f}(\vec{x};\theta) - y)\right)^2\right] = \left(E_\theta[\hat{f}(\vec{x};\theta)] - y\right)^2 + V_\theta[\hat{f}(\vec{x};\theta)]
\end{equation}
where $V_\theta$ is variance component and we have used our assumption that $y$, a test point label, has no noise. Based on Equation \ref{eq:bias-var-decomp}, a hypothesis is drawn where supposedly $\delta y(y) = H(x=y_t)\times\mathcal{N}(\mu,\sigma)$ is a zero-mean Gaussian distribution \cite{boonprong2018classification}, then the $\theta$ in the sensitive region would tend towards zero bias but high variance (as set by $\delta y(y)$). If $\delta \vec{x}(y)$ is a zero-mean Gaussian distribution, the bias would not tend towards zero because of regression dilution/attenuation \cite{frost2000correcting}. Instead the bias decreases to flattening $\hat{f}(\vec{x};\theta) = 0$. The variance also increases with $\delta\vec{x}(y)$.

Based on this analysis, we expect adding noise in $x$ will increase model bias and variance in the sensitive region. Adding noise to $y$ alone should not affect bias, but increase variance. This analysis is based on the assumption about the separability of $\theta$ into $s=1/s=0$ regions and may not hold. Hence, our results below explore this empirically in increasingly complex models.

\section{Methods}\label{sect:methods}

We apply and assess selective noise injection to data points that is deemed `sensitive' by adding zero-mean Gaussian noise to numerical features and labels. First, we define a sensitivity threshold $y_t$, which partitions the dataset into sensitive ($s(y_i) = 1$) and non-sensitive ($s(y_i) = 0$) data. Initially, this threshold is fixed at zero. In later experiments, $y_t$ is dynamically set based on the sensitivity split parameter $\alpha$. The value of $\alpha$ determines the fraction of data considered sensitive. Specifically, $y_t$ is the label value that separates the top-$\alpha$ fraction of the dataset. For example, if $\alpha=0.1$, the top 10\% of data points with the highest labels are classified as sensitive, and $y_t$ is the label value separating them from the rest. 

Once $y_t$ is determined, label noise $\delta y(y)$ or feature noise $\delta \vec{x}(y)$ is selectively applied to sensitive data in both training and validation sets. The standard deviations of zero-mean Gaussian noise, referred as ``noise level,'' control the extent of selective perturbation. We evaluate the model accuracy on unseen raw sensitive data points, as well as on non-sensitive data, to assess how well the model generalizes across both regions.

To add feature noise to selected SMILES representations, we replace the molecules with structurally similar ones using the ``superfast traversal, optimization, novelty, exploration, and discovery'' (STONED) method \cite{nigam2021beyond} with local chemical space generation as described by Wellawatte et al. \cite{wellawatte_seshadri_white_2021}. 
The similarity of the replacement molecule to the original molecule is measured by the Tanimoto similarity and ultimately determined by the noise level. All SMILES representations are then converted to molecular graphs for GCN training. Further details on SMILES feature noise can be found in SI. Model setup and training specifications also can be found in SI.

\section{Results}

To evaluate the effectiveness of selective noise in censoring sensitive data, we assess the model performance on three increasingly complex tasks: 1) polynomial regression on 1D synthetic data, 2) a multilayer perceptron (MLP) on high-dimensional synthetic data, and 3) a graph convolutional network (GCN) \cite{kipf2017semi} on an experimental molecular dataset to predict lipophilicity. For each task, we assess the performance numerically using Equation~\ref{eq:goal} and visually with a parity plot ($y$ vs. $\hat{f}(x)$) on unseen raw data.  

\subsection{Visualizing Selective Noise with 1D Polynomial Regression}

\begin{figure}
    \centering
    \captionsetup{width=0.9\linewidth}
    \includegraphics[width=\textwidth]{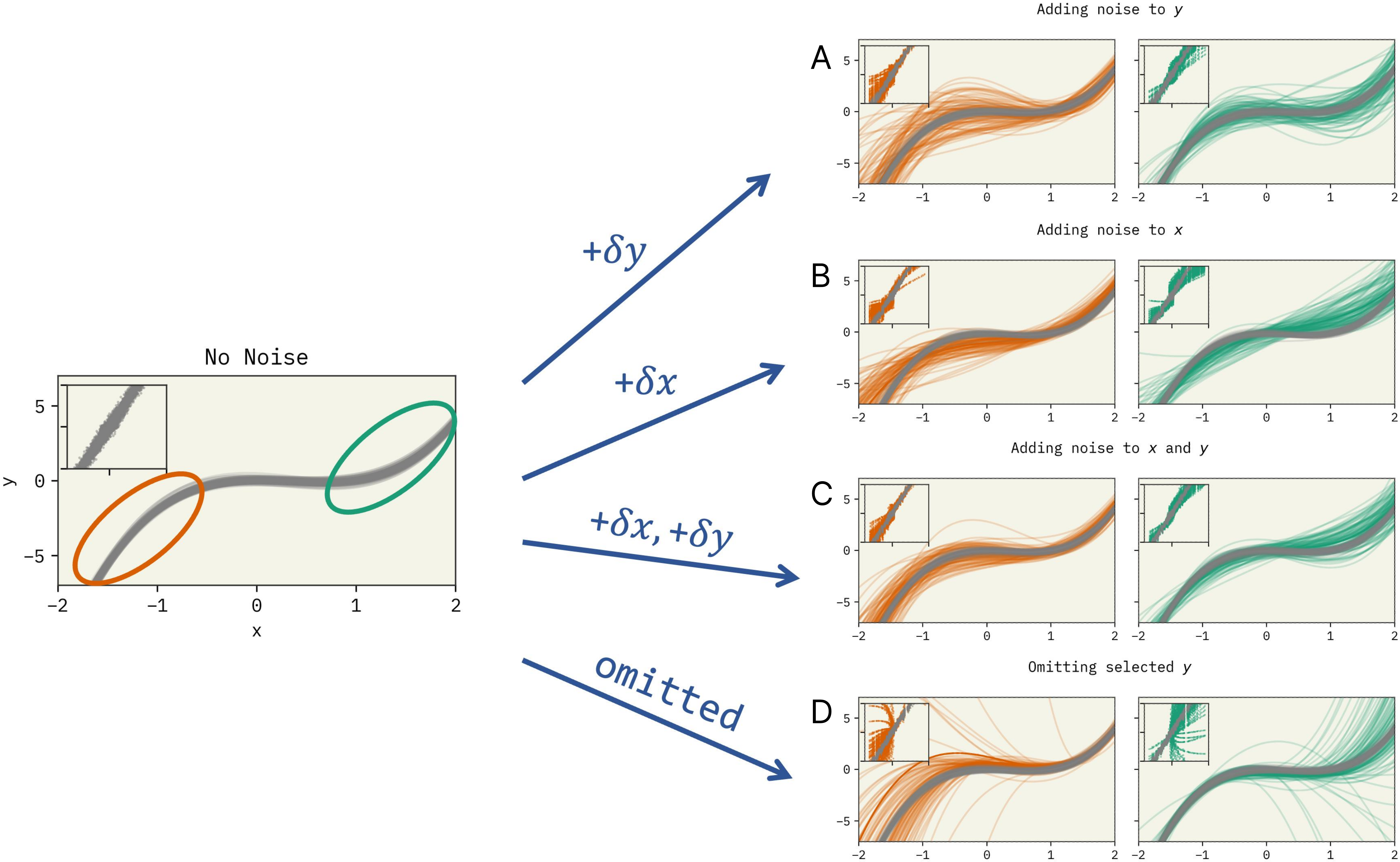}
    \caption{Fitted cubic curves after applying four different training data perturbations for 100 trials. 
    \textbf{Left}: The gray curve represents ground truth model ($x,y$).  
    \textbf{Right}: Models trained on perturbed data. Four perturbation types are applied in the following order: label noise ($\delta y(y)$) (\textbf{subplot A}), feature noise ($\delta x(y)$) (\textbf{subplot B}), and combined label \& feature noise ($\delta y(y)/\delta x(y)$) \textbf{(subplot C)}, and selective omission \textbf{(subplot D)}.
    Selective perturbation is applied to either of two data regions: data points with negative labels $y<0$ (orange) and data points with positive labels $y>0$ (green). \textbf{Inset} figure in each subplot shows a corresponding parity plot on raw unseen data.}
    \label{fig:polyreg}
\end{figure}

The nature of polynomial regression with 1D feature input allows us to visualize the fitted model on the $(x_i,y_i)$ plot and clearly interpret the differences in the fitting parameters after adding selective noise. Using a synthetic dataset, we applied polynomial regression with least squares to analyze fitted models under three conditions: raw data, noisy data (with three variations of selective noise), and filtered data after omission. Data perturbations, either selective noise or omission, are applied to two different regions of data: negatively labeled data (fitted curves in orange) and positively labeled data (green), as shown in Figure \textbf{\ref{fig:polyreg}}. For reference, curves $\hat{f}(x)$ fitted to true data $\mathcal{D}$ are shown in gray.

Original data $\mathcal{D}$ was generated from a cubic equation $y = x^3-x^2 + \sigma$, where $\sigma \sim \mathcal{N}(0,0.5)$ mimics the small inherent noise. Sensitivity was defined as $s(y)=1$ for data points where $y > 0$ or $y < 0$, as we studied both cases separately. Modified data $\mathcal{D}'$ was generated by adding Gaussian noise: $\delta x(y) \sim \mathcal{N}(0,0.5)$, $\delta y(y) \sim \mathcal{N}(0,5)$, or their combination $\delta x(y) +\delta y(y)$ in $s(y)=1$. For combined noise, standard deviations were halved. A cubic fit $\hat{f}(x)$ was applied to $\mathcal{D}'$ across 100 trials. Feature noise $\delta x$ had a greater impact than the label noise $\delta y$ for the same Gaussian noise level, requiring a higher standard deviation for $\delta y$. In the baseline omission method, removed data points were not replaced, leading to fewer data points for analysis and reflecting the nature of filtered data.

As seen in both fitted curves and parity plots in Figure \textbf{\ref{fig:polyreg}A}, adding selective label noise $\delta y$ to training data increases variance, particularly in the sensitive region. Selective feature noise $\delta x$ induces high bias in the sensitive region, evident from the suppression of curvature in fitted curves and the regional shifts in parity plots. Feature noise also introduces some variance, predominantly in the sensitive region. Selective noise thus allows controlled fitting to specific $y$ regions in polynomial regression, aligning with theoretical expectations in section \ref{sect:theory}. Applying both label and feature noise leads to increased variance and bias in the $s=1$ region. Attenuation error in parity plots further demonstrates that selective noise can specifically control bias at extreme values. 
The presence of attenuation error in the parity plots (inset figures in Figure \textbf{\ref{fig:polyreg}B} and \textbf{C}) further confirms that selective noise can selectively influence bias at extreme values.

Omitting data points where $y$ falls in the sensitive region creates a strong differential in generalization error. In one-dimensional polynomial regression on $(x_i,y_i)$, the model does not extrapolate beyond the training range, enabling the omission to censor sensitive $y$ values. However, we anticipate that deep learning models, such as MLPs, will be able to extrapolate well in high-dimensional settings, making omission less effective.

\subsection{Selective Noise in Multilayer Perceptrons (MLP) with Synthetic Data}

\begin{figure}
    \centering
    \captionsetup{width=0.9\linewidth}
    \includegraphics[width=0.9\textwidth]{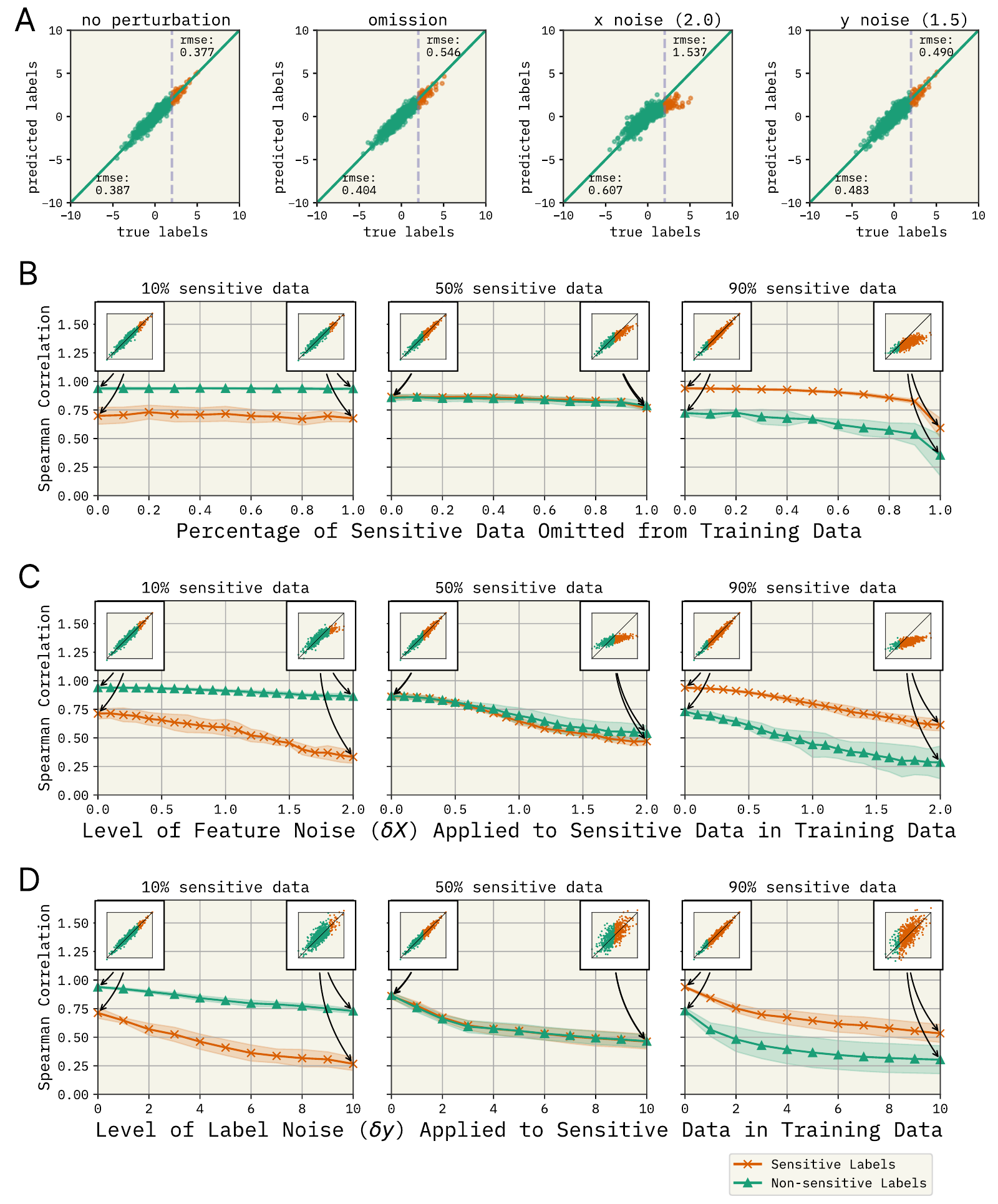}
    \caption{Evaluation of MLP Performance with Selective Noise. 
    \textbf{A.} A single instance of MLP parity plot after training on each type of data perturbation. We use $y>2$ here as the threshold for sensitive labels. Noise levels are $n_x=2.0$ for features and $n_y=1.5$ for labels.
    \textbf{B.} Spearman correlation of $y$ vs. $\hat{f}(x)$ on unseen raw data after training on selectively omitted data, dependent on the amount of sensitive data in the dataset.
    \textbf{C.} Spearman correlation on unseen raw data after training on selective feature noise.
    \textbf{D.} Spearman correlation on unseen raw data after training on selective label noise.
    }
    \label{fig:mlp}
\end{figure}

We used a multilayer perceptron (MLP) to evaluate a multidimensional case. To generate high-dimensional synthetic data, we initialized an MLP with random parameters and used it to create raw data $\mathcal{D}$ consisting of $(\vec{x_i},y_i)$ with feature dimension of 50. To maximize error in the sensitive region $s(y)=1$ while minimizing error in the non-sensitive region $s(y)=0$, we tested various levels of selective noise by adjusting the standard deviation of zero-mean Gaussian noise. For omission, we controlled the perturbation effect by varying the percentage of training data omitted in the sensitive region. In addition to using a fixed threshold $y > y_t$ (shown in Figure \textbf{\ref{fig:mlp}A}), we also analyzed different sensitive/non-sensitive splits. Figure \textbf{\ref{fig:mlp}B-D} presents the results for the 10\%, 50\%, and 90\% splits, showing the effect of noise level adjustments. Parity plots are included as insets to visualize the distribution of both sensitive and non-sensitive labels and to capture changes in distribution due to selective noise.

The baseline omission method has minimal impact on model accuracy, measured by the Spearman correlation of ground truth $y$ vs. prediction $\hat{f}(x)$. Varying the omission percentage provides little control over accuracy for both sensitive and non-sensitive labels. Omission is only effective when a large portion of the dataset is sensitive and  at least 80\% of these sensitive instances are removed from training and validation data. However, in such case (e.g., a 90\% sensitivity split), model accuracy drops significantly -- by about 35\% lower for sensitive labels and 50\% lower for non-sensitive label -- due to the small size of the remaining dataset.

Selective feature noise here is defined as $\delta \vec{x} = \mathcal{N}(\vec{0},\vec{1} \times n_x)$, where $n_x$ denotes the noise level. Figure \textbf{\ref{fig:mlp}B} shows that as feature noise increases, the trade-off between selective noise effectiveness (test error for sensitive labels) and desired accuracy (test error for non-sensitive labels) diminishes. At high feature noise ($n_x=2.0$) with a 10\% sensitivity split, generalization error for non-sensitive labels remains nearly unchanged, while the Spearman correlation for sensitive labels drops from 0.71 to 0.33. With a 50\% sensitivity split, accuracy declines by 45\% for sensitive labels and 37\% for non-sensitive labels. At extreme sensitivity levels (90\%), the trade-off becomes more severe: a 30\% accuracy drop for sensitive labels vs. 60\% for non-sensitive labels. This effect is only observed in highly sensitive datasets. Inset parity plots in Figure \textbf{\ref{fig:mlp}C} reveal attenuation bias, with flattening within the sensitive region. Overall, feature noise provides greater control over accuracy across all sensitivity splits compared to omission.

Label noise is defined as $\delta y = \mathcal{N}(0, n_y)$, where $n_y$ is the noise level. Injecting selective label noise induces a trade-off between model accuracy in both regions, decreasing at roughly the same rate as label noise increases across nearly all sensitivity splits. Inset parity plots in Figure \textbf{\ref{fig:mlp}D} show that label noise introduces variance across both regions, despite being selectively applied only to sensitive labels in training data.

In summary, for MLP on high-dimensional data, omission has little effect on accuracy except in extremely sensitive datasets, label noise primarily induces variance across both regions, and feature noise causes desired regional attenuation bias in low-to-moderate sensitivity settings.

\subsection{Selective Noise in Graph Convolutional Network (GCN) with Experimental Data}

\begin{figure}
    \centering
    \captionsetup{width=0.9\linewidth}
    \includegraphics[width=0.9\textwidth]{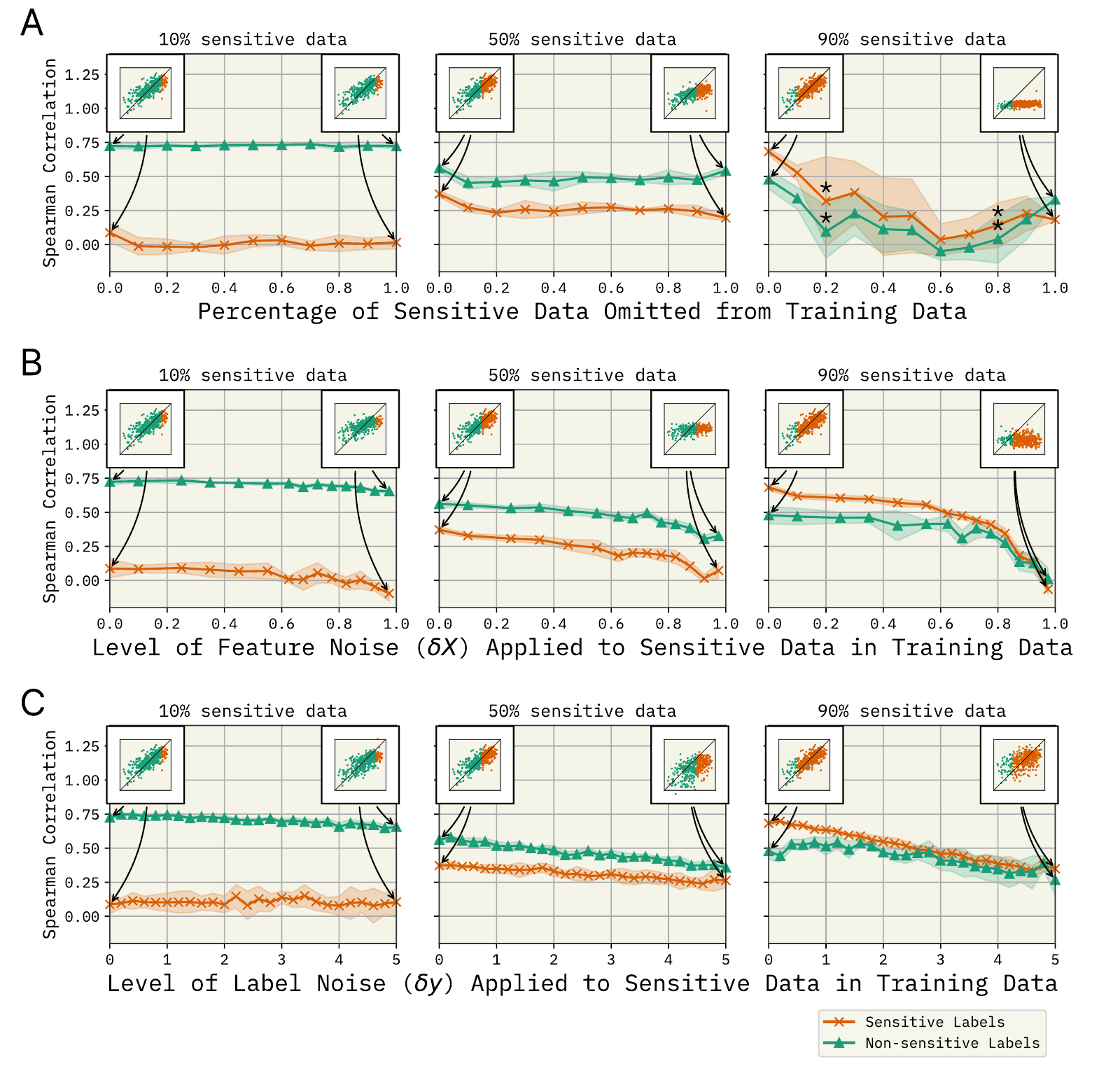}
    \caption{Evaluation of GCN Performance with Selective Noise. 
    \textbf{A.} Spearman correlation of $y$ vs. $\hat{f}(x)$ on unseen raw data after training on selectively omitted data, dependent on the amount of sensitive data in the dataset.
    \textbf{B.} Spearman correlation on unseen raw data after training on selective feature noise.
    \textbf{C.} Spearman correlation on unseen raw data after training on selective label noise.
    \textbf{*Note:} in 90\% sensitive data, some data failed to converge after omission, leading to missing trials.
    }
    \label{fig:gcn}
\end{figure}

As the final task, we evaluated censoring sensitive data using a graph convolutional network (GCN) trained with selective noise \cite{kipf2017semi}. Unlike previous tasks, we removed the assumption of no label noise and used experimental lipophilicity data, which measures a molecule’s preference for octanol over water \cite{wu2018moleculenet}. The dataset $\mathcal{D}$ consists of molecular graphs $(x)$ and measured lipophilicity logD $(y_i)$. Feature noise cannot be applied as Gaussian noise due to the molecular graph structure. Instead, we replaced molecules with structurally similar ones, where noise level is inversely related to Tanimoto similarity, with higher noise producing more dissimilar molecules (see \ref{sect:methods} section). As in previous models, Gaussian noise was selectively added to labels. We used Spearman correlations to mitigate magnitude effects from the dataset’s uneven distribution and sparsity as observed in zero-noise parity plots of our results. However, this does not fully eliminate the influence of the underlying data distribution, resulting in varying initial correlation values across sensitivity splits in Figure \textbf{\ref{fig:gcn}}.

Omitting sensitive data does not reduce the accuracy of sensitive predictions in the sensitivity splits 10\% and 50\%, confirming that omission is ineffective for censoring in deep learning models, which extrapolate relatively well for regression tasks. At 90\% sensitivity, omission results vary widely as the impact of randomness in selecting data points outweighs the effect of the omission fraction, making the trends between omission and accuracy difficult to discern.
Additionally, some training instances repeatedly failed to converge (marked by asterisks in Figure \textbf{\ref{fig:gcn}A}). This suggests that when a large portion of the dataset is sensitive, the omission method becomes difficult to modulate.

Compared to omission, selective feature noise effectively reduces accuracy for sensitive predictions across all sensitivity splits (Figure \textbf{\ref{fig:gcn}B}). The accuracy drop is more pronounced for sensitive labels than for non-sensitive ones: at a 10\% sensitivity split, accuracy decreases by 210\% for sensitive labels but only 9\% for non-sensitive labels. At the 50\% split, the drop is 80\% vs. 42\%, and at the 90\% split, it is 109\% vs. 97\%. As sensitivity increases to 90\%, correlation in the $s=1$ region drops further, yielding the lowest accuracy in GCN experiments. This highlights feature noise as an potentially effective censoring method in deep learning. Similar to MLP, label noise induces variance across both regions (Figure \textbf{\ref{fig:gcn}C}). Its impact is minimal at a 10\% sensitivity split but increases at the 90\% split, where correlation for sensitive labels steadily declines, indicating some control over predictions.

With GCN, feature noise selectively reduces accuracy for sensitive data while having a smaller effect on non-sensitive predictions. Label noise induces variance rather than attenuation bias but can still impact extreme labels. These effects persist even in experimental datasets with inherent noise. Omission method is unreliable with increasing portion of sensitivity in the dataset.

\section{Discussion}

We have shown that adding selective noise to the sensitive region in one-dimensional, multi-dimensional, and experimental datasets (including actual molecular structures) induces variance and bias in machine learning models. More specifically, feature noise selectively controls systematic errors by increasing attenuation bias in deep learning models while lowering accuracy for sensitive labels more effectively than omission. This effect persists even with experimental data containing inherent noise, as demonstrated by the GCN trained on a perturbed Lipophilicity dataset, where accuracy on unseen sensitive data dropped by up to 210\%. While feature noise effectively reduces accuracy for predicting sensitive information more than for non-sensitive data (i.e., the beneficial region), further improvements are needed to minimize unintended effects on non-sensitive predictions.

We also found that label noise increases model variance in the sensitive region. Unlike feature noise, which induces bias that persists across models, label noise effects may diminish in an ensemble setting due to variance reduction through averaging. Crucially, our study demonstrates that simply omitting sensitive data does not prevent extrapolation in deep learning models such as MLP and GCN, making omission an ineffective method for reducing predictive accuracy on sensitive chemical information.

In this work, we split data into two categories: sensitive and non-sensitive labels. However, real-world applications, such as toxicological datasets, may involve multiple sensitivity levels or risk categories. Future work could explore strategies for handling these nuanced classifications, investigating the risks of unmasking data perturbed with selective noise, improving accuracy preservation in the non-sensitive beneficial region, and extending feature noise to molecular-related representations like atom-in-SMILES tokenization \cite{ucak2023ais} and chemical reaction datasets.

\section*{Conclusion}

The selective noise approach provides a way to censor sensitive data, helping to mitigate dual-use risks in predictive chemistry (DURPC). This study investigates how adding selective noise can influence model performance, with the goal of reducing accuracy in sensitive regions while preserving it elsewhere for legitimate purposes. As discussed in the Theory section, label noise directly controls variance but does not affect bias in the sensitive region, while possibly increasing some variance in the non-sensitive region.
Meanwhile, feature noise provides greater control over which regions experience accuracy reduction and higher systematic error, with its effect becoming more pronounced as the proportion of sensitive information in the dataset grows. 

We have demonstrated that filtering out sensitive information alone does not prevent extrapolation in deep learning models for regression tasks, making omission ineffective for reducing model accuracy in sensitive chemical spaces. Our findings suggest that introducing feature noise during training may provide a more targeted way to influence model behavior in sensitive regions, though further research is needed to refine this approach.

While this work does not fully resolve dual-use concerns in machine learning for chemistry, it demonstrates an initial step toward understanding how data perturbations affect model performance and build responsible machine learning models. We hope this work will open the possibilities for future DURPC research work and promote safe open-source data sharing for academic research. 

\section*{Acknowledgments}
Research reported in this work was supported by the National Institute of General Medical Sciences of the National Institutes of Health under award number R35GM137966 and the University of Rochester CTSA award number UL1 TR002001 from the National Center for Advancing Translational Sciences of the National Institutes of Health. The authors thank the Center for Integrated Research Computing (CIRC) at the University of Rochester for providing computational resources and technical support. The content is solely the responsibility of the authors and does not necessarily represent the official views of the National Institutes of Health.

\section*{Code Availability}
The code is available at \url{https://github.com/ur-whitelab/chem-dual-use}.

\bibliographystyle{unsrtnat}
\bibliography{dual-use}

\begin{thebibliography}{93}
\providecommand{\natexlab}[1]{#1}
\providecommand{\url}[1]{\texttt{#1}}
\expandafter\ifx\csname urlstyle\endcsname\relax
  \providecommand{\doi}[1]{doi: #1}\else
  \providecommand{\doi}{doi: \begingroup \urlstyle{rm}\Url}\fi

\bibitem[Butler et~al.(2018)Butler, Davies, Cartwright, Isayev, and
  Walsh]{butler2018machine}
Keith~T Butler, Daniel~W Davies, Hugh Cartwright, Olexandr Isayev, and Aron
  Walsh.
\newblock Machine learning for molecular and materials science.
\newblock \emph{Nature}, 559\penalty0 (7715):\penalty0 547--555, 2018.

\bibitem[Moosavi et~al.(2020)Moosavi, Jablonka, and Smit]{moosavi2020role}
Seyed~Mohamad Moosavi, Kevin~Maik Jablonka, and Berend Smit.
\newblock The role of machine learning in the understanding and design of
  materials.
\newblock \emph{Journal of the American Chemical Society}, 142\penalty0
  (48):\penalty0 20273--20287, 2020.
\newblock \doi{10.1021/jacs.0c09105}.
\newblock URL \url{https://doi.org/10.1021/jacs.0c09105}.
\newblock PMID: 33170678.

\bibitem[Kim et~al.(2021)Kim, Park, Min, and Kim]{kim2021comprehensive}
Jintae Kim, Sera Park, Dongbo Min, and Wankyu Kim.
\newblock Comprehensive survey of recent drug discovery using deep learning.
\newblock \emph{International Journal of Molecular Sciences}, 22\penalty0
  (18):\penalty0 9983, 2021.

\bibitem[Ramos et~al.(2025)Ramos, Collison, and White]{ramos2025review}
Mayk~Caldas Ramos, Christopher~J Collison, and Andrew~D White.
\newblock A review of large language models and autonomous agents in chemistry.
\newblock \emph{Chemical Science}, 2025.

\bibitem[{U.S. Government}(2014)]{u.s._government_united_2014}
{U.S. Government}.
\newblock United {States} {Government} {Policy} for {Institutional} {Oversight}
  of {Life} {Sciences} {Dual} {Use} {Research} of {Concern}.
\newblock Technical report, United States Government, Washington, D.C.,
  September 2014.

\bibitem[Miller and Selgelid(2008)]{miller_ethical_2008}
Seumas Miller and Michael~J. Selgelid.
\newblock \emph{Ethical and {Philosophical} {Consideration} of the {Dual}-{Use}
  {Dilemma} in the {Biological} {Sciences}}.
\newblock Springer Science \& Business Media, July 2008.
\newblock ISBN 978-1-4020-8312-9.

\bibitem[Evans(2013)]{evans_great_2013}
Nicholas~G. Evans.
\newblock Great expectations—ethics, avian flu and the value of progress.
\newblock \emph{Journal of Medical Ethics}, 39\penalty0 (4):\penalty0 209--213,
  April 2013.
\newblock ISSN , 1473-4257.
\newblock \doi{10.1136/medethics-2012-100712}.
\newblock URL \url{http://jme.bmj.com/content/39/4/209}.

\bibitem[Lipsitch and Galvani(2014)]{lipsitch_ethical_2014}
Marc Lipsitch and Alison~P. Galvani.
\newblock Ethical {Alternatives} to {Experiments} with {Novel} {Potential}
  {Pandemic} {Pathogens}.
\newblock \emph{PLOS Medicine}, 11\penalty0 (5):\penalty0 e1001646, May 2014.
\newblock \doi{10.1371/journal.pmed.1001646}.
\newblock URL \url{http://dx.doi.org/10.1371/journal.pmed.1001646}.

\bibitem[Selgelid(2016)]{selgelid_gain--function_2016}
Michael~J. Selgelid.
\newblock Gain-of-{Function} {Research}: {Ethical} {Analysis}.
\newblock \emph{Science and Engineering Ethics}, 22\penalty0 (4):\penalty0
  923--964, August 2016.
\newblock ISSN 1353-3452, 1471-5546.
\newblock \doi{10.1007/s11948-016-9810-1}.
\newblock URL
  \url{https://link.springer.com/article/10.1007/s11948-016-9810-1}.

\bibitem[Koblentz(2017)]{koblentz_novo_2017}
Gregory~D. Koblentz.
\newblock The {De} {Novo} {Synthesis} of {Horsepox} {Virus}: {Implications} for
  {Biosecurity} and {Recommendations} for {Preventing} the {Reemergence} of
  {Smallpox}.
\newblock \emph{Health Security}, 15\penalty0 (6):\penalty0 620--628, 2017.
\newblock ISSN 2326-5108.
\newblock \doi{10.1089/hs.2017.0061}.

\bibitem[Bostrom(2014)]{bostrom_superintelligence:_2014}
Nick Bostrom.
\newblock \emph{Superintelligence: {Paths}, {Dangers}, {Strategies}}.
\newblock OUP Oxford, Oxford, July 2014.
\newblock ISBN 978-0-19-967811-2.

\bibitem[{UNCCCW}(2016)]{uncccw_report_2016}
{UNCCCW}.
\newblock Report of the 2016 {Informal} {Meeting} of {Experts} on {Lethal}
  {Autonomous} {Weapons} {Systems} ({LAWS}).
\newblock Technical report, United Nations, Geneva, 2016.

\bibitem[Tucker(2012)]{tucker_innovation_2012}
Jonathan~B. Tucker, editor.
\newblock \emph{Innovation, {Dual} {Use}, and {Security}: {Managing} the
  {Risks} of {Emerging} {Biological} and {Chemical} {Technologies}}.
\newblock MIT Press, Cambridge, MA, March 2012.
\newblock \doi{10.7551/mitpress/9147.001.0001}.
\newblock URL
  \url{https://direct.mit.edu/books/book/4418/Innovation-Dual-Use-and-SecurityManaging-the-Risks}.

\bibitem[Mehlich(2018)]{mehlich_chemistry_2018}
Jan Mehlich.
\newblock Chemistry and {Dual} {Use}: {From} {Scientific} {Integrity} to
  {Social} {Responsibility}.
\newblock \emph{Helvetica Chimica Acta}, 101\penalty0 (9), 2018.
\newblock ISSN 1522-2675.
\newblock \doi{10.1002/hlca.201800098}.
\newblock URL
  \url{https://onlinelibrary.wiley.com/doi/abs/10.1002/hlca.201800098}.
\newblock \_eprint:
  https://onlinelibrary.wiley.com/doi/pdf/10.1002/hlca.201800098.

\bibitem[noa(2021)]{noauthor_chemical_nodate}
Chemical {Facility} {Anti}-{Terrorism} {Standards} ({CFATS}), 2021.
\newblock URL
  \url{https://www.ecfr.gov/current/title-6/chapter-I/part-27?toc=1}.

\bibitem[{Organization for the Prohibition of Chemical
  Weapons}(1998)]{organization_for_the_prohibition_of_chemical_weapons_convention_1998}
{Organization for the Prohibition of Chemical Weapons}.
\newblock The {Convention} on the {Prohibition} of the {Development},
  {Production}, {Stockpiling} and {Use} of {Chemical} {Weapons} and on their
  {Destruction}: {Annex} on {Chemicals}, 1998.
\newblock URL \url{https://www.opcw.org/chemical-weapons-convention}.

\bibitem[Rein et~al.(2023)Rein, Hou, Stickland, Petty, Pang, Dirani, Michael,
  and Bowman]{rein2023gpqa}
David Rein, Betty~Li Hou, Asa~Cooper Stickland, Jackson Petty, Richard~Yuanzhe
  Pang, Julien Dirani, Julian Michael, and Samuel~R Bowman.
\newblock {GPQA}: A graduate-level google-proof {Q}\&{A} benchmark.
\newblock \emph{arXiv preprint arXiv:2311.12022}, 2023.

\bibitem[White et~al.(2023)White, Hocky, Gandhi, Ansari, Cox, Wellawatte,
  Sasmal, Yang, Liu, Singh, et~al.]{white2023assessment}
Andrew~D White, Glen~M Hocky, Heta~A Gandhi, Mehrad Ansari, Sam Cox, Geemi~P
  Wellawatte, Subarna Sasmal, Ziyue Yang, Kangxin Liu, Yuvraj Singh, et~al.
\newblock Assessment of chemistry knowledge in large language models that
  generate code.
\newblock \emph{Digital Discovery}, 2\penalty0 (2):\penalty0 368--376, 2023.

\bibitem[Bran et~al.(2023)Bran, Cox, White, and Schwaller]{bran2023chemcrow}
Andres~M Bran, Sam Cox, Andrew~D White, and Philippe Schwaller.
\newblock Chemcrow: Augmenting large-language models with chemistry tools,
  2023.

\bibitem[Mirza et~al.(2024)Mirza, Alampara, Kunchapu, R{\'\i}os-Garc{\'\i}a,
  Emoekabu, Krishnan, Gupta, Schilling-Wilhelmi, Okereke, Aneesh,
  et~al.]{mirza2024large}
Adrian Mirza, Nawaf Alampara, Sreekanth Kunchapu, Marti{\~n}o
  R{\'\i}os-Garc{\'\i}a, Benedict Emoekabu, Aswanth Krishnan, Tanya Gupta, Mara
  Schilling-Wilhelmi, Macjonathan Okereke, Anagha Aneesh, et~al.
\newblock Are large language models superhuman chemists?
\newblock \emph{arXiv preprint arXiv:2404.01475}, 2024.

\bibitem[He et~al.(2023)He, Feng, Min, Yi, Tang, Li, Zhang, Chen, Zhou, Xie,
  et~al.]{he2023control}
Jiyan He, Weitao Feng, Yaosen Min, Jingwei Yi, Kunsheng Tang, Shuai Li, Jie
  Zhang, Kejiang Chen, Wenbo Zhou, Xing Xie, et~al.
\newblock Control risk for potential misuse of artificial intelligence in
  science.
\newblock \emph{arXiv preprint arXiv:2312.06632}, 2023.

\bibitem[Tang et~al.(2024)Tang, Jin, Zhu, Yuan, Zhang, Zhou, Qu, Zhao, Tang,
  Zhang, et~al.]{tang2024prioritizing}
Xiangru Tang, Qiao Jin, Kunlun Zhu, Tongxin Yuan, Yichi Zhang, Wangchunshu
  Zhou, Meng Qu, Yilun Zhao, Jian Tang, Zhuosheng Zhang, et~al.
\newblock Prioritizing safeguarding over autonomy: Risks of {LLM} agents for
  science.
\newblock \emph{arXiv preprint arXiv:2402.04247}, 2024.

\bibitem[Rose(2022)]{rose2022openai}
J~Rose.
\newblock {OpenAI}’s new chatbot will tell you how to shoplift and make
  explosives. {Vice}, 2022.
\newblock URL
  \url{https://www.vice.com/en/article/xgyp9j/openais-new-chatbot-will-tell-you-how-to-shoplift-and-make-explosives,}.

\bibitem[Boiko et~al.(2023)Boiko, MacKnight, and Gomes]{boiko2023emergent}
Daniil~A. Boiko, Robert MacKnight, and Gabe Gomes.
\newblock Emergent autonomous scientific research capabilities of large
  language models, 2023.

\bibitem[Qi et~al.(2023)Qi, Zeng, Xie, Chen, Jia, Mittal, and
  Henderson]{qi2023fine}
Xiangyu Qi, Yi~Zeng, Tinghao Xie, Pin-Yu Chen, Ruoxi Jia, Prateek Mittal, and
  Peter Henderson.
\newblock Fine-tuning aligned language models compromises safety, even when
  users do not intend to!
\newblock \emph{arXiv preprint arXiv:2310.03693}, 2023.

\bibitem[Gopal et~al.(2023)Gopal, Helm-Burger, Justen, Soice, Tzeng,
  Jeyapragasan, Grimm, Mueller, and Esvelt]{gopal2023will}
Anjali Gopal, Nathan Helm-Burger, Lenni Justen, Emily~H Soice, Tiffany Tzeng,
  Geetha Jeyapragasan, Simon Grimm, Benjamin Mueller, and Kevin~M Esvelt.
\newblock Will releasing the weights of large language models grant widespread
  access to pandemic agents?
\newblock \emph{arXiv preprint arXiv:2310.18233}, 2023.

\bibitem[Gong et~al.(2023)Gong, Ran, Liu, Wang, Cong, Wang, Duan, and
  Wang]{gong2023figstep}
Yichen Gong, Delong Ran, Jinyuan Liu, Conglei Wang, Tianshuo Cong, Anyu Wang,
  Sisi Duan, and Xiaoyun Wang.
\newblock Figstep: Jailbreaking large vision-language models via typographic
  visual prompts.
\newblock \emph{arXiv preprint arXiv:2311.05608}, 2023.

\bibitem[{Gryphon Scientific}(2015)]{gryphon_scientific_risk_2015}
{Gryphon Scientific}.
\newblock Risk and {Benefit} {Analysis} of {Gain} of {Function} {Research}.
\newblock Technical report, National Science Advisory Board for Biosecurity,
  Bethesda, MD, 2015.

\bibitem[{National Academies of Sciences, Engineering, and
  Medicine.}(2018)]{national_academies_of_sciences_engineering_and_medicine_biodefense_2018}
{National Academies of Sciences, Engineering, and Medicine.}
\newblock \emph{Biodefense in the {Age} of {Synthetic} {Biology}}.
\newblock National Academies Press, Washington, D.C., December 2018.
\newblock ISBN 978-0-309-46518-2.
\newblock \doi{10.17226/24890}.
\newblock URL \url{https://doi.org/10.17226/24890}.

\bibitem[Sandberg and Nelson(2020)]{sandberg_who_2020}
Anders Sandberg and Cassidy Nelson.
\newblock Who should we fear more: biohackers, disgruntled postdocs, or bad
  governments? {A} simple risk chain model of biorisk.
\newblock \emph{Health Security}, 18\penalty0 (3):\penalty0 155--163, June
  2020.
\newblock ISSN 2326-5094.
\newblock \doi{10.1089/hs.2019.0115}.
\newblock URL \url{https://www.ncbi.nlm.nih.gov/pmc/articles/PMC7310205/}.

\bibitem[Cameron(1999)]{cameron_multi-track_1999}
Gavin Cameron.
\newblock Multi-track {Microproliferation}: {Lessons} from {Aum} {Shinrikyo}
  and {Al} {Qaida}.
\newblock \emph{Studies in Conflict \& Terrorism}, 22\penalty0 (4):\penalty0
  277--309, November 1999.
\newblock ISSN 1057-610X.
\newblock \doi{10.1080/105761099265658}.
\newblock URL
  \url{https://www.tandfonline.com/doi/citedby/10.1080/105761099265658}.
\newblock Publisher: Routledge.

\bibitem[Kuhlau et~al.(2011)Kuhlau, Höglund, Evers, and
  Eriksson]{kuhlau_precautionary_2011}
Frida Kuhlau, Anna~T. Höglund, Kathinka Evers, and Stefan Eriksson.
\newblock A {Precautionary} {Principle} for {Dual} {Use} {Research} in the
  {Life} {Sciences}.
\newblock \emph{Bioethics}, 25\penalty0 (1):\penalty0 1--8, 2011.
\newblock ISSN 1467-8519.
\newblock \doi{10.1111/j.1467-8519.2009.01740.x}.
\newblock URL
  \url{https://onlinelibrary.wiley.com/doi/abs/10.1111/j.1467-8519.2009.01740.x}.
\newblock \_eprint:
  https://onlinelibrary.wiley.com/doi/pdf/10.1111/j.1467-8519.2009.01740.x.

\bibitem[Clarke(2013)]{clarke_precautionary_2013}
Steve Clarke.
\newblock The {Precautionary} {Principle} and the {Dual}-{Use} {Dilemma}.
\newblock In Brian Rappert and Michael~J. Selgelid, editors, \emph{On the
  {Dual} {Uses} of {Science} and {Ethics}}, Principles, {Practices}, and
  {Prospects}, pages 223--234. ANU Press, 2013.
\newblock ISBN 978-1-925021-33-2.
\newblock URL \url{https://www.jstor.org/stable/j.ctt5hgz15.19}.

\bibitem[Yuan et~al.(2024)Yuan, Li, Wang, Chen, Mao, Huang, Xue, Wang, Ren, and
  Wang]{yuan2024seval}
Xiaohan Yuan, Jinfeng Li, Dongxia Wang, Yuefeng Chen, Xiaofeng Mao, Longtao
  Huang, Hui Xue, Wenhai Wang, Kui Ren, and Jingyi Wang.
\newblock S-eval: Automatic and adaptive test generation for benchmarking
  safety evaluation of large language models.
\newblock \emph{arXiv preprint arXiv:2405.14191}, 2024.

\bibitem[Phuong et~al.(2024)Phuong, Aitchison, Catt, Cogan, Kaskasoli,
  Krakovna, Lindner, Rahtz, Assael, Hodkinson, et~al.]{phuong2024evaluating}
Mary Phuong, Matthew Aitchison, Elliot Catt, Sarah Cogan, Alexandre Kaskasoli,
  Victoria Krakovna, David Lindner, Matthew Rahtz, Yannis Assael, Sarah
  Hodkinson, et~al.
\newblock Evaluating frontier models for dangerous capabilities.
\newblock \emph{arXiv preprint arXiv:2403.13793}, 2024.

\bibitem[Tedeschi et~al.(2024)Tedeschi, Friedrich, Schramowski, Kersting,
  Navigli, Nguyen, and Li]{tedeschi2024alert}
Simone Tedeschi, Felix Friedrich, Patrick Schramowski, Kristian Kersting,
  Roberto Navigli, Huu Nguyen, and Bo~Li.
\newblock {ALERT}: A comprehensive benchmark for assessing large language
  models' safety through red teaming.
\newblock \emph{arXiv preprint arXiv:2404.08676}, 2024.

\bibitem[Vidgen et~al.(2024)Vidgen, Agrawal, Ahmed, Akinwande, Al-Nuaimi,
  Alfaraj, Alhajjar, Aroyo, Bavalatti, Blili-Hamelin,
  et~al.]{vidgen2024introducing}
Bertie Vidgen, Adarsh Agrawal, Ahmed~M Ahmed, Victor Akinwande, Namir
  Al-Nuaimi, Najla Alfaraj, Elie Alhajjar, Lora Aroyo, Trupti Bavalatti,
  Borhane Blili-Hamelin, et~al.
\newblock Introducing v0.5 of the {AI} safety benchmark from {MLC}ommons.
\newblock \emph{arXiv preprint arXiv:2404.12241}, 2024.

\bibitem[Rebedea et~al.(2023)Rebedea, Dinu, Sreedhar, Parisien, and
  Cohen]{rebedea-etal-2023-nemo}
Traian Rebedea, Razvan Dinu, Makesh~Narsimhan Sreedhar, Christopher Parisien,
  and Jonathan Cohen.
\newblock {N}e{M}o guardrails: A toolkit for controllable and safe {LLM}
  applications with programmable rails.
\newblock In Yansong Feng and Els Lefever, editors, \emph{Proceedings of the
  2023 Conference on Empirical Methods in Natural Language Processing: System
  Demonstrations}, pages 431--445, Singapore, December 2023. Association for
  Computational Linguistics.
\newblock \doi{10.18653/v1/2023.emnlp-demo.40}.
\newblock URL \url{https://aclanthology.org/2023.emnlp-demo.40}.

\bibitem[Inan et~al.(2023)Inan, Upasani, Chi, Rungta, Iyer, Mao, Tontchev, Hu,
  Fuller, Testuggine, et~al.]{inan2023llama}
Hakan Inan, Kartikeya Upasani, Jianfeng Chi, Rashi Rungta, Krithika Iyer,
  Yuning Mao, Michael Tontchev, Qing Hu, Brian Fuller, Davide Testuggine,
  et~al.
\newblock Llama {G}uard: {LLM}-based input-output safeguard for human-{AI}
  conversations.
\newblock \emph{arXiv preprint arXiv:2312.06674}, 2023.

\bibitem[Grattafiori et~al.(2024)Grattafiori, Dubey, Jauhri, Pandey, Kadian,
  et~al.]{grattafiori2024llama3}
Aaron Grattafiori, Abhimanyu Dubey, Abhinav Jauhri, Abhinav Pandey, Abhishek
  Kadian, et~al.
\newblock The {L}lama 3 herd of models.
\newblock \emph{arXiv}, 2407.21783, 2024.

\bibitem[Challacombe and Haas(2024)]{challacombe2024towards}
Chance~A Challacombe and Nikhil~S Haas.
\newblock Towards a dataset for state of the art protein toxin classification.
\newblock \emph{bioRxiv}, pages 2024--04, 2024.

\bibitem[McMahan et~al.(2017)McMahan, Moore, Ramage, Hampson, and
  y~Arcas]{mcmahan2017communication}
Brendan McMahan, Eider Moore, Daniel Ramage, Seth Hampson, and Blaise~Aguera
  y~Arcas.
\newblock Communication-efficient learning of deep networks from decentralized
  data.
\newblock In \emph{Artificial intelligence and statistics}, pages 1273--1282.
  {PMLR}, 2017.

\bibitem[Yang et~al.(2019)Yang, Liu, Chen, and Tong]{yang2019federated}
Qiang Yang, Yang Liu, Tianjian Chen, and Yongxin Tong.
\newblock Federated machine learning: Concept and applications.
\newblock \emph{{ACM} Transactions on Intelligent Systems and Technology
  {(TIST)}}, 10\penalty0 (2):\penalty0 1--19, 2019.

\bibitem[Chen et~al.(2020)Chen, Xue, Chuai, Yang, and Liu]{chen2020fl}
Shaoqi Chen, Dongyu Xue, Guohui Chuai, Qiang Yang, and Qi~Liu.
\newblock {FL-QSAR}: a federated learning-based {QSAR} prototype for
  collaborative drug discovery.
\newblock \emph{Bioinformatics}, 36\penalty0 (22-23):\penalty0 5492--5498,
  2020.

\bibitem[Heyndrickx et~al.(2023)Heyndrickx, Mervin, Morawietz, Sturm,
  Friedrich, Zalewski, Pentina, Humbeck, Oldenhof, Niwayama,
  et~al.]{heyndrickx2023melloddy}
Wouter Heyndrickx, Lewis Mervin, Tobias Morawietz, No{\'e} Sturm, Lukas
  Friedrich, Adam Zalewski, Anastasia Pentina, Lina Humbeck, Martijn Oldenhof,
  Ritsuya Niwayama, et~al.
\newblock {MELLODDY}: {C}ross-pharma federated learning at unprecedented scale
  unlocks benefits in qsar without compromising proprietary information.
\newblock \emph{Journal of {C}hemical {I}nformation and {M}odeling},
  64\penalty0 (7):\penalty0 2331--2344, 2023.

\bibitem[Karr et~al.(2005)Karr, Feng, Lin, Sanil, Young, and
  Reiter]{karr2005secure}
Alan~F Karr, Jun Feng, Xiaodong Lin, Ashish~P Sanil, S~Stanley Young, and
  Jerome~P Reiter.
\newblock Secure analysis of distributed chemical databases without data
  integration.
\newblock \emph{Journal of {C}omputer-{A}ided {M}olecular {D}esign},
  19:\penalty0 739--747, 2005.

\bibitem[Chennakesavalu et~al.(2024)Chennakesavalu, Hu, Ibarraran, and
  Rotskoff]{chennakesavalu2024energy}
Shriram Chennakesavalu, Frank Hu, Sebastian Ibarraran, and Grant~M Rotskoff.
\newblock Energy rank alignment: Using preference optimization to search
  chemical space at scale.
\newblock \emph{arXiv preprint arXiv:2405.12961}, 2024.

\bibitem[Ouyang et~al.(2022)Ouyang, Wu, Jiang, Almeida, Wainwright, Mishkin,
  Zhang, Agarwal, Slama, Ray, et~al.]{ouyang2022training}
Long Ouyang, Jeffrey Wu, Xu~Jiang, Diogo Almeida, Carroll Wainwright, Pamela
  Mishkin, Chong Zhang, Sandhini Agarwal, Katarina Slama, Alex Ray, et~al.
\newblock Training language models to follow instructions with human feedback.
\newblock \emph{Advances in neural information processing systems},
  35:\penalty0 27730--27744, 2022.

\bibitem[Bai et~al.(2022)Bai, Jones, Ndousse, Askell, Chen, DasSarma, Drain,
  Fort, Ganguli, Henighan, et~al.]{bai2022training}
Yuntao Bai, Andy Jones, Kamal Ndousse, Amanda Askell, Anna Chen, Nova DasSarma,
  Dawn Drain, Stanislav Fort, Deep Ganguli, Tom Henighan, et~al.
\newblock Training a helpful and harmless assistant with reinforcement learning
  from human feedback.
\newblock \emph{arXiv preprint arXiv:2204.05862}, 2022.

\bibitem[Saisubramanian et~al.(2022)Saisubramanian, Zilberstein, and
  Kamar]{saisubramanian2022avoiding}
Sandhya Saisubramanian, Shlomo Zilberstein, and Ece Kamar.
\newblock Avoiding negative side effects due to incomplete knowledge of {AI}
  systems.
\newblock \emph{AI Magazine}, 42\penalty0 (4):\penalty0 62--71, 2022.

\bibitem[Zhang et~al.(2018)Zhang, Durfee, and Singh]{zhang2018minimax}
Shun Zhang, Edmund~H Durfee, and Satinder Singh.
\newblock Minimax-regret querying on side effects for safe optimality in
  factored markov decision processes.
\newblock In \emph{IJCAI}, pages 4867--4873, 2018.

\bibitem[Low and Kumar(2024)]{low2024safe}
Siow~Meng Low and Akshat Kumar.
\newblock Safe reinforcement learning with learned non-markovian safety
  constraints.
\newblock \emph{arXiv preprint arXiv:2405.03005}, 2024.

\bibitem[Stooke et~al.(2020)Stooke, Achiam, and Abbeel]{stooke2020responsive}
Adam Stooke, Joshua Achiam, and Pieter Abbeel.
\newblock Responsive safety in reinforcement learning by {PID} {L}agrangian
  methods.
\newblock In \emph{International Conference on Machine Learning}, pages
  9133--9143. PMLR, 2020.

\bibitem[Ganguli et~al.(2022)Ganguli, Lovitt, Kernion, Askell, Bai, Kadavath,
  Mann, Perez, Schiefer, Ndousse, et~al.]{ganguli2022red}
Deep Ganguli, Liane Lovitt, Jackson Kernion, Amanda Askell, Yuntao Bai, Saurav
  Kadavath, Ben Mann, Ethan Perez, Nicholas Schiefer, Kamal Ndousse, et~al.
\newblock Red teaming language models to reduce harms: Methods, scaling
  behaviors, and lessons learned.
\newblock \emph{arXiv preprint arXiv:2209.07858}, 2022.

\bibitem[Mitchell et~al.(2022)Mitchell, Henderson, Manning, Jurafsky, and
  Finn]{mitchell2022selfdestructing}
Eric Mitchell, Peter Henderson, Christopher~D. Manning, Dan Jurafsky, and
  Chelsea Finn.
\newblock Self-destructing models: Increasing the costs of harmful dual uses in
  foundation models, 2022.

\bibitem[Rosati et~al.(2024)Rosati, Wehner, Williams, Łukasz Bartoszcze,
  Atanasov, Gonzales, Majumdar, Maple, Sajjad, and
  Rudzicz]{rosati2024representation}
Domenic Rosati, Jan Wehner, Kai Williams, Łukasz Bartoszcze, David Atanasov,
  Robie Gonzales, Subhabrata Majumdar, Carsten Maple, Hassan Sajjad, and Frank
  Rudzicz.
\newblock Representation noising: A defence mechanism against harmful
  finetuning, 2024.
\newblock URL \url{https://arxiv.org/abs/2405.14577}.

\bibitem[Kirchenbauer et~al.(2023)Kirchenbauer, Geiping, Wen, Katz, Miers, and
  Goldstein]{kirchenbauer2023watermark}
John Kirchenbauer, Jonas Geiping, Yuxin Wen, Jonathan Katz, Ian Miers, and Tom
  Goldstein.
\newblock A watermark for large language models, 2023.

\bibitem[Urbina et~al.(2022)Urbina, Lentzos, Invernizzi, and
  Ekins]{urbina_dual_2022}
Fabio Urbina, Filippa Lentzos, Cédric Invernizzi, and Sean Ekins.
\newblock Dual use of artificial-intelligence-powered drug discovery.
\newblock \emph{Nature Machine Intelligence}, 4\penalty0 (3):\penalty0
  189--191, March 2022.
\newblock ISSN 2522-5839.
\newblock \doi{10.1038/s42256-022-00465-9}.
\newblock URL \url{https://www.nature.com/articles/s42256-022-00465-9}.
\newblock Number: 3 Publisher: Nature Publishing Group.

\bibitem[Filimonov and Poroikov(2005)]{filimonov2005relevant}
Dmitry Filimonov and Vladimir Poroikov.
\newblock Why relevant chemical information cannot be exchanged without
  disclosing structures.
\newblock \emph{Journal of computer-aided molecular design}, 19:\penalty0
  705--713, 2005.

\bibitem[Masek et~al.(2008)Masek, Shen, Smith, and Pearlman]{masek2008sharing}
Brian~B Masek, Lingling Shen, Karl~M Smith, and Robert~S Pearlman.
\newblock Sharing chemical information without sharing chemical structure.
\newblock \emph{Journal of chemical information and modeling}, 48\penalty0
  (2):\penalty0 256--261, 2008.

\bibitem[Faulon et~al.(2005)Faulon, Brown, and Martin]{faulon2005reverse}
Jean-Loup Faulon, W~Michael Brown, and Shawn Martin.
\newblock Reverse engineering chemical structures from molecular descriptors:
  how many solutions?
\newblock \emph{Journal of computer-aided molecular design}, 19:\penalty0
  637--650, 2005.

\bibitem[Balaban(2005)]{balaban2005can}
Alexandru~T Balaban.
\newblock Can topological indices transmit information on properties but not on
  structures?
\newblock \emph{Journal of computer-aided molecular design}, 19:\penalty0
  651--660, 2005.

\bibitem[Varnek et~al.(2005)Varnek, Fourches, Hoonakker, and
  Solov’ev]{varnek2005substructural}
Alexandre Varnek, Denis Fourches, Frank Hoonakker, and Vitaly~P Solov’ev.
\newblock Substructural fragments: an universal language to encode reactions,
  molecular and supramolecular structures.
\newblock \emph{Journal of computer-aided molecular design}, 19:\penalty0
  693--703, 2005.

\bibitem[Nicholls and Grant(2005)]{nicholls2005molecular}
Anthony Nicholls and J~Andrew Grant.
\newblock Molecular shape and electrostatics in the encoding of relevant
  chemical information.
\newblock \emph{Journal of computer-aided molecular design}, 19:\penalty0
  661--686, 2005.

\bibitem[Matlock and Swamidass(2014)]{matlock2014sharing}
Matthew Matlock and S~Joshua Swamidass.
\newblock Sharing chemical relationships does not reveal structures.
\newblock \emph{Journal of Chemical Information and Modeling}, 54\penalty0
  (1):\penalty0 37--48, 2014.

\bibitem[Bologa et~al.(2005)Bologa, Allu, Olah, Kappler, and
  Oprea]{bologa2005descriptor}
Cristian Bologa, Tharun~Kumar Allu, Marius Olah, Michael~A Kappler, and Tudor~I
  Oprea.
\newblock Descriptor collision and confusion: Toward the design of descriptors
  to mask chemical structures.
\newblock \emph{Journal of Computer-Aided Molecular Design}, 19:\penalty0
  625--635, 2005.
\newblock \doi{https://doi.org/10.1007/s10822-005-9020-4}.

\bibitem[Tetko et~al.(2005)Tetko, Abagyan, and Oprea]{tetko2005surrogate}
Igor~V Tetko, Ruben Abagyan, and Tudor~I Oprea.
\newblock Surrogate data--a secure way to share corporate data.
\newblock \emph{Journal of computer-aided molecular design}, 19:\penalty0
  749--764, 2005.
\newblock \doi{https://doi.org/10.1007/s10822-005-9013-3}.

\bibitem[Bradley(2005)]{bradley2005share}
David Bradley.
\newblock Share and share alike.
\newblock \emph{Nature Reviews Drug Discovery}, 4\penalty0 (3):\penalty0
  180--180, March 2005.
\newblock ISSN 1474-1784.
\newblock \doi{10.1038/nrd1683}.
\newblock URL \url{https://doi.org/10.1038/nrd1683}.

\bibitem[Trepalin and Osadchiy(2005)]{trepalin2005centroidal}
Sergey Trepalin and Nikolay Osadchiy.
\newblock The centroidal algorithm in molecular similarity and diversity
  calculations on confidential datasets.
\newblock \emph{Journal of computer-aided molecular design}, 19:\penalty0
  715--729, 2005.

\bibitem[Kaiser et~al.(2005)Kaiser, Zdrazil, and Ecker]{kaiser2005similarity}
Dominik Kaiser, Barbara Zdrazil, and Gerhard~F Ecker.
\newblock Similarity-based descriptors {(SIBAR)}--{A} tool for safe exchange of
  chemical information?
\newblock \emph{Journal of computer-aided molecular design}, 19:\penalty0
  687--692, 2005.

\bibitem[Clement and G{\"u}ner(2005)]{clement2005possibilities}
Omoshile~O Clement and Osman~F G{\"u}ner.
\newblock Possibilities for transfer of relevant data without revealing
  structural information.
\newblock \emph{Journal of computer-aided molecular design}, 19\penalty0
  (9):\penalty0 731--738, 2005.

\bibitem[Swamidass et~al.(2015)Swamidass, Matlock, and
  Rozenblit]{swamidass2015securely}
S~Joshua Swamidass, Matthew Matlock, and Leon Rozenblit.
\newblock Securely measuring the overlap between private datasets with
  cryptosets.
\newblock \emph{PloS one}, 10\penalty0 (2):\penalty0 e0117898, 2015.

\bibitem[Mouton et~al.(2024)Mouton, Lucas, and Guest]{mouton2024operational}
Christopher~A Mouton, Caleb Lucas, and Ella Guest.
\newblock \emph{The Operational Risks of {AI} in Large-Scale Biological
  Attacks: Results of a Red-Team Study}.
\newblock RAND, 2024.
\newblock URL \url{https://www.rand.org/pubs/research_reports/RRA2977-2.html}.

\bibitem[Patwardhan et~al.(2024)Patwardhan, Liu, Markov, Chowdhury, Leet, Cone,
  Maltbie, Huizinga, Wainwright, Jackson, Adler, Casagrande, and
  Madry]{patwardhan2024building}
Tejal Patwardhan, Kevin Liu, Todor Markov, Neil Chowdhury, Dillon Leet, Natalie
  Cone, Caitlin Maltbie, Joost Huizinga, Carroll Wainwright, Shawn~(Froggi)
  Jackson, Steven Adler, Rocco Casagrande, and Aleksander Madry.
\newblock Building an early warning system for {LLM}-aided biological threat
  creation.
\newblock
  \url{https://openai.com/research/building-an-early-warning-system-for-llm-aided-biological-threat-creation},
  2024.

\bibitem[Ge et~al.(2023)Ge, Zhou, Hou, Khabsa, Wang, Wang, Han, and
  Mao]{ge2023mart}
Suyu Ge, Chunting Zhou, Rui Hou, Madian Khabsa, Yi-Chia Wang, Qifan Wang,
  Jiawei Han, and Yuning Mao.
\newblock Mart: Improving {LLM} safety with multi-round automatic red-teaming.
\newblock \emph{arXiv preprint arXiv:2311.07689}, 2023.

\bibitem[Li et~al.(2024{\natexlab{a}})Li, Li, Yin, Ahmed, Liu, and
  Liu]{li2024red}
Mukai Li, Lei Li, Yuwei Yin, Masood Ahmed, Zhenguang Liu, and Qi~Liu.
\newblock Red teaming visual language models.
\newblock \emph{arXiv preprint arXiv:2401.12915}, 2024{\natexlab{a}}.

\bibitem[Li et~al.(2024{\natexlab{b}})Li, Pan, Gopal, Yue, Berrios, Gatti, Li,
  Dombrowski, Goel, Phan, et~al.]{li2024wmdp}
Nathaniel Li, Alexander Pan, Anjali Gopal, Summer Yue, Daniel Berrios, Alice
  Gatti, Justin~D Li, Ann-Kathrin Dombrowski, Shashwat Goel, Long Phan, et~al.
\newblock The {WMDP} benchmark: Measuring and reducing malicious use with
  unlearning.
\newblock \emph{arXiv preprint arXiv:2403.03218}, 2024{\natexlab{b}}.

\bibitem[Barrett et~al.(2024)Barrett, Jackson, Murphy, Madkour, and
  Newman]{barrett2024benchmark}
Anthony~M Barrett, Krystal Jackson, Evan~R Murphy, Nada Madkour, and Jessica
  Newman.
\newblock Benchmark early and red team often: A framework for assessing and
  managing dual-use hazards of {AI} foundation models.
\newblock \emph{arXiv preprint arXiv:2405.10986}, 2024.

\bibitem[Xu et~al.(2023)Xu, Ji, Li, and Lu]{xu2023small}
Pengcheng Xu, Xiaobo Ji, Minjie Li, and Wencong Lu.
\newblock Small data machine learning in materials science.
\newblock \emph{NPJ Computational Materials}, 9\penalty0 (1):\penalty0 42,
  2023.

\bibitem[Altae-Tran et~al.(2017)Altae-Tran, Ramsundar, Pappu, and
  Pande]{altae2017low}
Han Altae-Tran, Bharath Ramsundar, Aneesh~S Pappu, and Vijay Pande.
\newblock Low data drug discovery with one-shot learning.
\newblock \emph{ACS Central Science}, 3\penalty0 (4):\penalty0 283--293, 2017.

\bibitem[Wang et~al.(2018)Wang, Girshick, Hebert, and
  Hariharan]{wang2018lowshot}
Yu-Xiong Wang, Ross Girshick, Martial Hebert, and Bharath Hariharan.
\newblock Low-shot learning from imaginary data, 2018.

\bibitem[Boonprong et~al.(2018)Boonprong, Cao, Chen, Ni, Xu, and
  Acharya]{boonprong2018classification}
Sornkitja Boonprong, Chunxiang Cao, Wei Chen, Xiliang Ni, Min Xu, and
  Bipin~Kumar Acharya.
\newblock The classification of noise-afflicted remotely sensed data using
  three machine-learning techniques: Effect of different levels and types of
  noise on accuracy.
\newblock \emph{ISPRS International Journal of Geo-Information}, 7\penalty0
  (7), 2018.
\newblock ISSN 2220-9964.
\newblock \doi{10.3390/ijgi7070274}.
\newblock URL \url{https://www.mdpi.com/2220-9964/7/7/274}.

\bibitem[Adlam and Pennington(2020)]{adlam2020understanding}
Ben Adlam and Jeffrey Pennington.
\newblock Understanding double descent requires a fine-grained bias-variance
  decomposition.
\newblock \emph{Advances in Neural Information Processing Systems},
  33:\penalty0 11022--11032, 2020.

\bibitem[Geman et~al.(1992)Geman, Bienenstock, and Doursat]{geman1992neural}
Stuart Geman, Elie Bienenstock, and Ren{\'e} Doursat.
\newblock Neural networks and the bias/variance dilemma.
\newblock \emph{Neural Computation}, 4\penalty0 (1):\penalty0 1--58, 1992.

\bibitem[Frost and Thompson(2000)]{frost2000correcting}
Chris Frost and Simon~G Thompson.
\newblock Correcting for regression dilution bias: comparison of methods for a
  single predictor variable.
\newblock \emph{Journal of the Royal Statistical Society: Series A (Statistics
  in Society)}, 163\penalty0 (2):\penalty0 173--189, 2000.

\bibitem[Nigam et~al.(2021)Nigam, Pollice, Krenn, dos Passos~Gomes, and
  Aspuru-Guzik]{nigam2021beyond}
AkshatKumar Nigam, Robert Pollice, Mario Krenn, Gabriel dos Passos~Gomes, and
  Alan Aspuru-Guzik.
\newblock Beyond generative models: superfast traversal, optimization, novelty,
  exploration and discovery {(STONED) }algorithm for molecules using {SELFIES}.
\newblock \emph{Chemical science}, 12\penalty0 (20):\penalty0 7079--7090, 2021.

\bibitem[Wellawatte et~al.(2022)Wellawatte, Seshadri, and
  White]{wellawatte_seshadri_white_2021}
Geemi~P Wellawatte, Aditi Seshadri, and Andrew~D White.
\newblock Model agnostic generation of counterfactual explanations for
  molecules.
\newblock \emph{Chemical Science}, 13\penalty0 (13):\penalty0 3697--3705, 2022.

\bibitem[Kipf and Welling(2017)]{kipf2017semi}
Thomas~N. Kipf and Max Welling.
\newblock Semi-supervised classification with graph convolutional networks,
  2017.

\bibitem[Wu et~al.(2018)Wu, Ramsundar, Feinberg, Gomes, Geniesse, Pappu,
  Leswing, and Pande]{wu2018moleculenet}
Zhenqin Wu, Bharath Ramsundar, Evan~N Feinberg, Joseph Gomes, Caleb Geniesse,
  Aneesh~S Pappu, Karl Leswing, and Vijay Pande.
\newblock {MoleculeNet}: a benchmark for molecular machine learning.
\newblock \emph{Chemical Science}, 9\penalty0 (2):\penalty0 513--530, 2018.

\bibitem[Ucak et~al.(2023)Ucak, Ashyrmamatov, and Lee]{ucak2023ais}
Umit~V. Ucak, Islambek Ashyrmamatov, and Juyong Lee.
\newblock {Improving the quality of chemical language model outcomes with
  atom-in-SMILES tokenization}.
\newblock \emph{Journal of Cheminformatics}, 15\penalty0 (1):\penalty0 55,
  2023.
\newblock \doi{10.1186/s13321-023-00725-9}.

\bibitem[Rogers and Hahn(2010)]{Rogers2010ECFP}
David Rogers and Mathew Hahn.
\newblock Extended-connectivity fingerprints.
\newblock \emph{Journal of Chemical Information and Modeling}, 50\penalty0
  (5):\penalty0 742--754, 2010.
\newblock \doi{10.1021/ci100050t}.
\newblock URL \url{https://pubs.acs.org/doi/10.1021/ci100050t}.

\bibitem[Li et~al.(2021)Li, Zhou, Hu, Fan, Zhang, Gu, and Karypis]{dgllife}
Mufei Li, Jinjing Zhou, Jiajing Hu, Wenxuan Fan, Yangkang Zhang, Yaxin Gu, and
  George Karypis.
\newblock {DGL-LifeSci}: An open-source toolkit for deep learning on graphs in
  life science.
\newblock \emph{ACS Omega}, 6\penalty0 (41):\penalty0 27233--27238, 2021.

\bibitem[Wang et~al.(2020)Wang, Zheng, Ye, Gan, Li, Song, Zhou, Ma, Yu, Gai,
  Xiao, He, Karypis, Li, and Zhang]{wang2020dgl}
Minjie Wang, Da~Zheng, Zihao Ye, Quan Gan, Mufei Li, Xiang Song, Jinjing Zhou,
  Chao Ma, Lingfan Yu, Yu~Gai, Tianjun Xiao, Tong He, George Karypis, Jinyang
  Li, and Zheng Zhang.
\newblock Deep {Graph} {Library}: A graph-centric, highly-performant package
  for graph neural networks, 2020.

\end{thebibliography}

\appendix

\section{Supplemental Information}

\subsection{Perturbing SMILES For GCN Feature Noise}\label{si:gen_smiles}
\begin{table}[h]
\centering
\captionsetup{width=0.6\linewidth, justification=centering}
\caption{Mapping between feature noise level and Tanimoto similarity scores for the GCN task}
\begin{tabular}{cc}
\toprule
SMILES Noise Level & Tanimoto Similarity Score Range \\ \midrule
0 & 1.0 \\
0.1 & 0.8 -1.0 \\
0.25 & 0.7 - 0.8 \\
0.35 & 0.6 - 0.7 \\
0.45 & 0.5 - 0.6 \\
0.55 & 0.4 - 0.5 \\
0.625 & 0.35 - 0.4 \\
0.675 & 0.3 - 0.35 \\
0.725 & 0.25 - 0.3 \\
0.775 & 0.2 - 0.25 \\
0.825 & 0.15 - 0.2 \\
0.875 & 0.1 - 0.15 \\
0.925 & 0.05-0.1 \\
0.975 & 0 - 0.05 \\ \bottomrule
\end{tabular}
\end{table}

To introduce feature noise to SMILES data, the original SMILES string is replaced by a new one, with the noise level controlled by selecting a range of Tanimoto similarity scores. Candidate molecules are generated by ``mutating'' the original molecule and selecting the first sample that falls within the desired similarity range. We sample similar candidate molecules using the ``superfast traversal, optimization, novelty, exploration and discovery'' (STONED) \cite{nigam2021beyond} with local chemical space generation as described by Wellawatte et al. \cite{wellawatte_seshadri_white_2021}. The ``noise level'' is the certain range of maximum distance we allow between the true molecule and the generated molecule, called Tanimoto similarity score. The Tanimoto similarity is computed using binary Extended-Connectivity Fingerprints (ECFP) with radius of 2 \cite{Rogers2010ECFP}.

We applied a uniform algorithm to set STONED parameters that generate candidate molecules for all noise levels. Initially, we set \verb|max_mutations| = 1, \verb|min_mutations| = 1, and a pool size of \verb|num_samples| = 15. If no suitable molecule is found within the target Tanimoto similarity range, the sample size increases by 10 for each iteration until a suitable candidate is found. The parameter \verb|max_mutations| also increases by 1 every iteration for low Tanimoto scores (<0.3).

\begin{figure}[ht]
  \centering
  \captionsetup{width=0.9\linewidth}
  \subcaptionbox{\label{smi_fig:a}}
  {\includegraphics[width=0.2\linewidth]{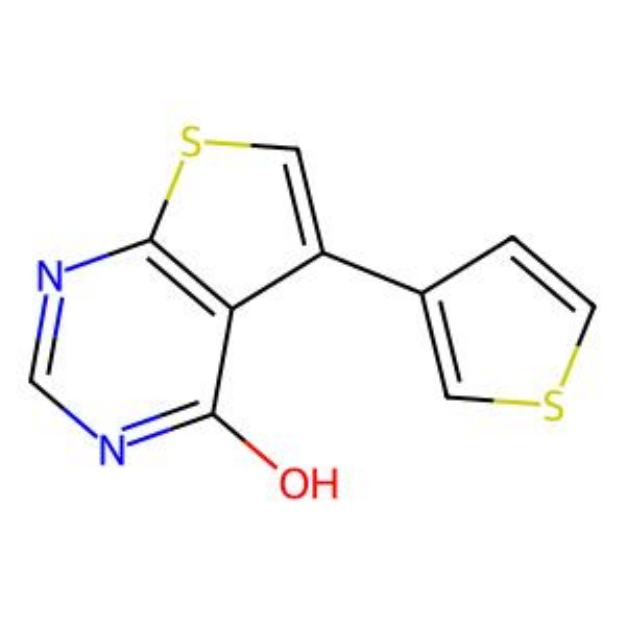}}
  \subcaptionbox{\label{smi_fig:b}}
  {\includegraphics[width=0.2\linewidth]{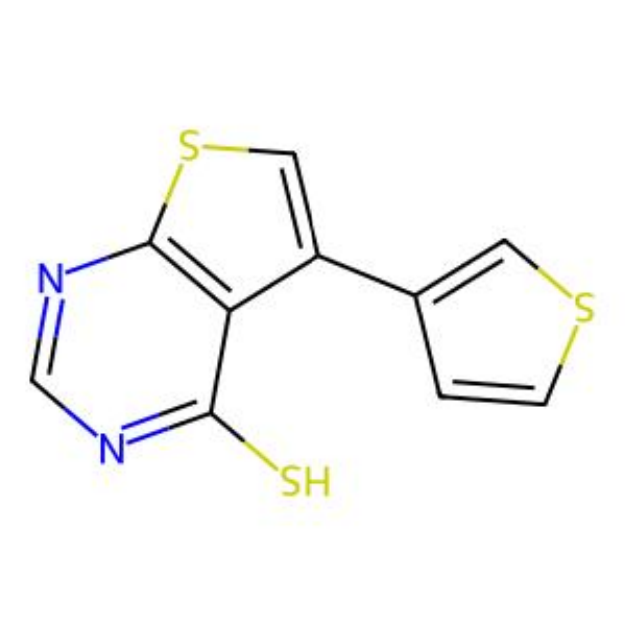}}
  \caption{The left is the original SMILES found in Lipophilicity dataset \cite{wu2018moleculenet}. 
  The right is closest similar molecule generated by STONED method \cite{wellawatte_seshadri_white_2021}, 
  with Tanimoto score of 0.76.}
  \label{fig:low_noise_smiles}
\end{figure}

If no suitable candidate is found after 10 iterations of pool generation, the original data point is removed. This occurred only at the lowest noise level (Tanimoto score between 0.8 and 1.0), where we had to remove about 10\% of the sensitive data points from the Lipophilicity dataset. This is primarily due to the difficulty in replacing small molecules. Figure \ref{fig:low_noise_smiles} demonstrates one example of how small molecule structures are highly sensitive to even tiny changes. Wellawatte et al. showed that highly similar molecules have a sparser distribution compared to molecules with low Tanimoto similarity, with fewer mutations needed for the latter \cite{wellawatte_seshadri_white_2021}.

\subsection{Model \& Training specifications}\label{si:training-specs}

\begin{table}[ht]
\centering
\caption{Hyperparameters for Deep Learning Models}
\begin{tabular}{l|l|l}
\hline
Hyperparameter & MLP & GCN \\
\hline
Dimension of Features (\(D_{in}\)) & 50 & n/a \\
Learning Rate (\(lr\)) & 0.001 & 0.005 \\
Hidden Dimension & 64 & 128 \\
Batch Size & 32 & 5 \\
Data Size & 6400 & 4200 \\
Split (Test/Val/Train) & 10/10/80\% & 10/10/80\% \\
Epochs & 60 & 60 \\
Patience for Early Stopping & 5 & 5 \\
Minimum Delta (\(min\_delta\)) for Early Stopping & \(1 \times 10^{-4}\) & \(1 \times 10^{-4}\) \\
\hline
\end{tabular}
\label{table:model_config}
\end{table}

The hyperparameters used during model training for multilayer perceptrons (MLP) and graph convolution network (GCN) are listed in Table \ref{table:model_config}. All parity plots in this paper were plotted using `original' data, i.e. no noise addition.

To evaluate the effectiveness of selective noise in censoring sensitive data, we assess model performance on three tasks: 1) polynomial regression with 1D synthetic data, 2) multilayer perceptron (MLP) on high-dimensional synthetic data, and 3) a graph convolutional network (GCN) on an experimental molecular dataset to predict lipophilicity. For each task, we measure the performance of applying selective noise numerically using Equation~\ref{eq:goal} and visualize results via parity plots ($y$ vs. $\hat{f}(x)$). %

\paragraph{1-D Polynomial Regression} We performed 1D polynomial regression on synthetic data $\mathcal{D}=\{(x_i, y_i)\}$. 200 data points were generated from the cubic equation $y = x^3-x^2 + \sigma$, where $\sigma \sim \mathcal{N}(0,0.5)$ to mimic inherent noise. 
For each of 100 trials, we randomly sampled 25 points to form a training set. The sensitive data were defined using a threshold $y_t =0$, classifying the points with either $y > 0$ or $y < 0$ as in the sensitive region ($s(y)=1$). We applied three types of selective noise: feature noise ($\delta x \sim \mathcal{N}(0,0.5)$), label noise ($\delta y \sim \mathcal{N}(0,5)$), and combined noise ($\delta x \sim \mathcal{N}(0,0.25)$, $\delta y \sim \mathcal{N}(0,2.5)$). Abaseline method omitted all sensitive data points and used only non-sensitive data for regression. The noisy training set then was fit to the cubic equation using least squares.

In the baseline method, which involves omission of data points, no re-selection is conducted to compensate for these omissions. Consequently, this typically results in the analysis involving fewer than 25 data points.

\paragraph{MLP} We applied selective noise to a synthetic dataset $\mathcal{D}=\{(\vec{x}_i, y_i)\}$ and trained a multilayer perceptron (MLP) on the noisy data. For each of five random seeds, we generated a dataset with 500 points, each with 50 input features, using a separate generative MLP. The sensitive region, $s(y)=1$, was defined using three different sensitivity splits: 10\%, 50\%, and 90\%. As described earlier, $y_t$ is the label value that separates the top-$\alpha$ fraction of the dataset, where $\alpha$ corresponds to the sensitivity split. Sensitive data points with $y > y_t$ in training and validation sets had Gaussian noise applied to their features and/or labels. Feature noise levels ranged from 0 and 2, and label noise levels ranged from 0 to 10. Both generative and predictive MLPs had 2 hidden layers and used a rectified linear unit (ReLU) activation function. The initial weights were randomized, differing between generative and predictive MLPs.

\paragraph{Graph Convolution Network}
We trained a graph convolution network (GCN) on the Lipophilicity dataset from MoleculeNet, originally curated from ChEMBL \cite{kipf2017semi,wu2018moleculenet}. We labeled the dataset into sensitive and non-sensitive regions based on logP thresholds of $y_t=3.51$, $2.33$, and $0.4$, for sensitivity splits at 10\%, 50\%, and 90\%, respectively. Similar to MLP study, we apply selective noise to training and validation data then measure model accuracy for predicting unseen data points with no noise added. Feature noise was added by replacing SMILES within a specified Tanimoto similarity range and label noise using zero-mean Gaussian noise. SMILES strings were then converted to molecular graphs using the canonical node featurizers from the DGL-LifeSci library, which featurizes using atom-related properties such as the atom type, atom degree, hydrogen atom counts, radical electrons, formal charge, hybridization, and aromaticity \cite{dgllife}. The GCN model is built with DGL (Deep Graph Library) \cite{wang2020dgl}, consisted of two graph convolution layers with same hidden dimensions for each. ReLU activation function was assigned after each graph convolution layer. After that, the readout is done by averaging the node features, and the dense layer as the final layer gives a scalar output. To isolate the effect of data perturbation despite high inherent noise in the dataset and model, the training, validation, and test sets are pre-split once and reused across all GCN experiments.

\end{document}